\documentclass{article}
\PassOptionsToPackage{numbers, compress}{natbib}
\usepackage[preprint]{neurips_2025}
\usepackage[utf8]{inputenc} %
\usepackage[T1]{fontenc}    %
\usepackage{hyperref}       %
\usepackage{url}            %
\usepackage{booktabs}       %
\usepackage{amsfonts}       %
\usepackage{nicefrac}       %
\usepackage{microtype}      %
\usepackage{xcolor}         %
\usepackage{amsmath}
\usepackage{multirow}
\usepackage{graphicx}  
\usepackage{enumitem}

\usepackage{subcaption}

\newcommand{\ourmodel}{\textbf{CAMS}}

\title{\ourmodel: A CityGPT-Powered Agentic Framework for Urban Human Mobility Simulation}

\author{
\textbf{Yuwei Du, Jie Feng, Jian Yuan, Yong Li} \\
Department of Electronic Engineering, BRNist, Tsinghua University, Beijing, China \\
\texttt{\{fengjie,liyong07\}@tsinghua.edu.cn}
}

\begin{document}

\maketitle

\begin{abstract}
Human mobility simulation plays a crucial role in various real-world applications. Recently, to address the limitations of traditional data-driven approaches, researchers have explored leveraging the commonsense knowledge and reasoning capabilities of large language models (LLMs) to accelerate human mobility simulation. However, these methods suffer from several critical shortcomings, including inadequate modeling of urban spaces and poor integration with both individual mobility patterns and collective mobility distributions. To address these challenges, we propose \textbf{C}ityGPT-Powered \textbf{A}gentic framework for \textbf{M}obility \textbf{S}imulation (\ourmodel), an agentic framework that leverages the language based urban foundation model to simulate human mobility in urban space. \ourmodel~comprises three core modules, including MobExtractor to extract template mobility patterns and synthesize new ones based on user profiles, GeoGenerator to generate anchor points considering collective knowledge and generate candidate urban geospatial knowledge using an enhanced version of CityGPT, TrajEnhancer to retrieve spatial knowledge based on mobility patterns and generate trajectories with real trajectory preference alignment via DPO. Experiments on real-world datasets show that \ourmodel~achieves superior performance without relying on externally provided geospatial information. Moreover, by holistically modeling both individual mobility patterns and collective mobility constraints, \ourmodel~generates more realistic and plausible trajectories. In general, \ourmodel~establishes a new paradigm that integrates the agentic framework with urban-knowledgeable LLMs for human mobility simulation. 
\end{abstract}

\section{Introduction}

Human mobility simulation is a critical real-world task with widespread applications across many domains~\cite{pappalardo2023future}, such as supporting the implementation of the 15-minute city concept in urban development by modeling residents' daily activities~\cite{zheng2023spatial}, optimizing transportation strategies through travel behavior simulation, and validating intervention policies in epidemic prevention and control. Given its significant value, the research community has studied this problem extensively for many years, resulting in a range of effective solutions. Early efforts, such as mechanism-based models like TimeGeo~\cite{jiang2016timegeo}, have gradually been supplemented—and surpassed—by recent deep learning approaches, such as MoveSim~\cite{feng2020learning}, ActSTD~\cite{yuan2022activity}, and DSTPP~\cite{yuan2023spatio} and on. Despite remarkable progress, key challenges remain—particularly concerning the spatial transferability of methods, as well as the controllability and interpretability of the generated mobility behaviors.

To address these challenges, recent research has explored integrating LLMs into mobility simulation, leveraging their role-playing~\cite{gao2024large, wang2024survey, gao2023s3, piao2025agentsociety}, commonsense knowledge~\cite{ChatGPT, touvron2023llama, ding2024understanding} and reasoning capabilities~\cite{wei2022emergent, xu2025towards} to achieve promising results. The most crucial and challenging aspect of applying LLMs to mobility simulation lies in effectively incorporating spatial information. Existing work~\cite{gurnee2023language, shao2024beyond} has shown that simply utilizing general-purpose LLMs is insufficient for accurately understanding urban space. As a result, studies such as CoPB~\cite{shao2024beyond} and LLM-Mob~\cite{wang2024simulating} have proposed specific mechanisms within their frameworks to mitigate this limitation and harness the strengths of LLMs for sequential modeling and reasoning. However, these approaches typically combine spatial knowledge and LLMs in a relatively independent manner, fusing information in an ad hoc fashion. Moreover, spatial knowledge is often simplified to facilitate model comprehension, and the integration process remains largely unidirectional, lacking feedback-driven optimization or iterative reasoning updates.

Recently, urban LLMs such as CityGPT~\cite{feng2024citygpt} and LAMP~\cite{balsebre2024lamp} have emerged, directly enhancing general LLMs with urban spatial knowledge through post-training and achieving impressive results on geospatial tasks such as urban spatial knowledge question answering. In these works, they convert the urban spatial knowledge into the language format and train the general model to enhance the urban spatial knowledge. This progress offers a new perspective on incorporating spatial knowledge into LLMs and enables deeper collaboration between spatial knowledge and spatial reasoning. 

\begin{figure*}[!t]
    \centering
    \includegraphics[width=\linewidth]{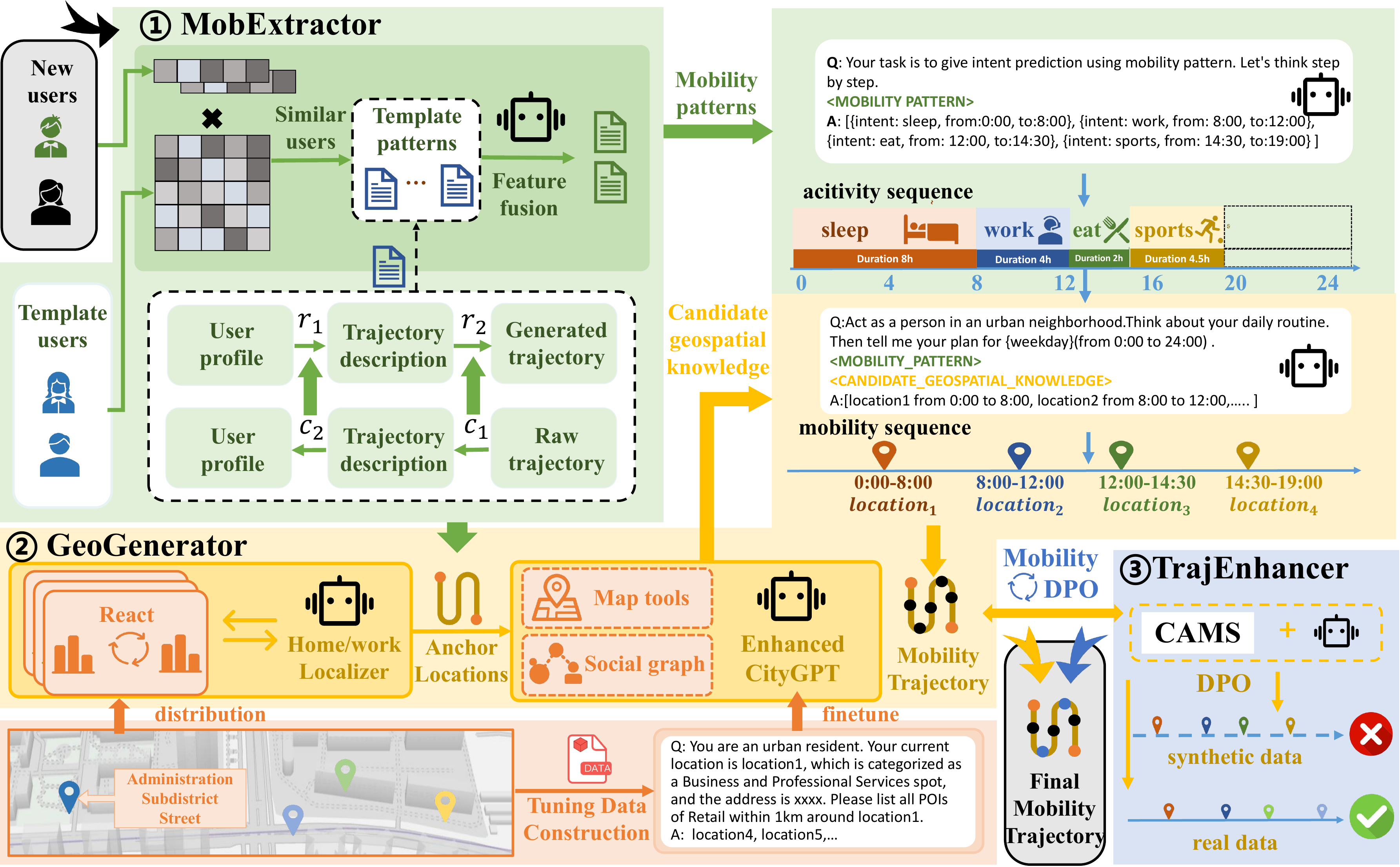}
    \caption{\ourmodel: \textbf{C}ityGPT-powered \textbf{A}gentic framework for urban human \textbf{M}obility \textbf{S}imulation, including MobExtractor, GeoGenerator and TrajEnhancer.}
    \label{fig:CAMS}
    \vspace{-20pt}
\end{figure*}

In this paper, we propose an agentic mobility simulation framework, \ourmodel, built upon CityGPT, by integrating native urban spatial knowledge into the reasoning process of large language models, enabling more controllable, accurate, and generalizable human mobility simulation. \ourmodel~comprises three core components that work in synergy to enable accurate and generalizable urban human mobility simulation. First, MobExtractor is designed to extract and summarize general mobility patterns from raw trajectory data, capturing diverse high-level behavioral regularities. Second, GeoGenerator leverages an enhanced version of CityGPT to generate synthetic mobility trajectories, using the activity sequence from MobExtractor as input, and incorporates rich geospatial knowledge into the mobility simulation process. Third, TrajEnhancer improves spatial-temporal consistency by aligning generated trajectories with real-world trajectory data through direct preference optimization, ensuring realism and coherence. Built upon this unified framework, multi-dimensional feedback mechanisms are naturally introduced to iteratively refine the mobility generation procedure, enhancing both the fidelity and adaptability of simulated human mobility.

In summary, our contributions are:
\begin{itemize}[leftmargin=1.5em,itemsep=0pt,parsep=0.2em,topsep=0.0em,partopsep=0.0em]
    \item To the best of our knowledge, this work introduces the first agentic framework that integrates a urban foundation model with rich geospatial knowledge and multi-dimensional feedback signals, embedding urban structure constraints into LLM reasoning for controllable and generalizable mobility simulation. 
    \item We propose a dual-phase architecture that first condenses template users' mobility patterns into compact linguistic representation, then generates synthetic patterns for new users through profile-aware feature fusion and variational encoding.
    \item Through geospatial information alignment and fine-grained urban geographic knowledge fine-tuning, we enhance CityGPT's capability to extract urban geospatial knowledge relevant to user profile and mobility patterns.
    \item Through iterated DPO training, we progressively enhance the spatiotemporal continuity of generated trajectories, strengthening the model's ability to capture the intrinsic connections between mobility patterns and urban geospatial knowledge.
    \item Experimental results on real-world datasets show that the proposed \ourmodel~framework, enhanced by incorporating an urban-knowledgeable large language model for geospatial reasoning and agentic simulation framework for mobility behavior reasoning, significantly outperforms existing methods in human mobility simulation. 
\end{itemize}

\section{Methodology}\label{headings}

In this paper, we propose \ourmodel, an agentic framework for generating trajectories in real urban spaces based on an urban-knowledgeable LLM, CityGPT~\cite{feng2024citygpt}. To align with the existing spatial knowledge in LLM effectively, we express urban structure in a hierarchical address system,  which is similar to human spatial cognition~\cite{feng2024agentmove}. In addition, we inject fine-grained urban mobility information into CityGPT to more thoroughly explore the information of each POI in urban space. 
The whole framework is shown in Figure~\ref{fig:CAMS}, which comprises three central components: MobExtractor, GeoGenerator and TrajEnhancer. First, we present MobExtractor in section~\ref{mob_generation}, which is designed to extract and synthesize mobility patterns in linguistic representations. Subsequently, we introduce the GeoGenerator capable of generating candidate urban geospatial knowledge related to user profile and mobility patterns in section~\ref{geogenerator}. Finally, we detail the TrajEnhancer in section~\ref{trajenhancer}, which generates trajectories in real urban spaces via integrated reasoning, and enhance trajectory generation with aligning the real-world preference.

\vspace{-5pt}
\subsection{MobExtractor: Semantic Mobility Pattern Extraction from Raw Trajectory} \label{sec:mobextractor}

MobExtractor employs a dual-phase architecture that first condenses template users' mobility patterns into compact linguistic representation, then generates synthetic patterns for new users through profile-aware feature fusion and variational encoding. User mobility patterns can be decomposed into shared generic patterns(common across populations) and special individual patterns(profile-specific variations). Data-driven approaches typically require massive high-quality trajectory datasets to effectively capture above mobility patterns, facing data scarcity problems. In contrast, LLM agents leverage their inherent knowledge to identify generic patterns by analyzing trajectories of a small set of users,  and synthesize individual patterns for other users through semantic profiling of user attributes. To enhance the model's capability to identify mobility patterns, we employ a two-step compression-recovery process in reconstruction stage. For test users, we employ an embedding-based method to synthesize movement patterns in generation stage.

\textbf{Mobility pattern recovery.} As shown in Figure \ref{fig:CAMS}, in the mobility patterns reconstruction phase, the model learns high-level correlations between user profiles, semantic trajectory descriptions, and raw mobility patterns through a dual-phase compression-reconstruction process. The model automatically distill observed patterns and correlations into interpretable natural language rules, including $\mathbf{c}_{1}$, $\mathbf{c}_{2}$ in compression stage and $\mathbf{r}_{1}$, $\mathbf{r}_{2}$ in reconstruction stage.
\begin{itemize}[leftmargin=1.5em,itemsep=0pt,parsep=0.2em,topsep=0.0em,partopsep=0.0em]
    \item  \textbf{Compression} In compression stage, the model learns compression patterns that map raw trajectory data to user profile representations, i.e., (1) How to derive users' behavioral habits and motivations by analyzing statistical patterns in their historical trajectories, (2) How to identify user's mobility pattern from raw trajectory, habits, motivations and address information, (3) How to identify profile-influencing features from trajectory descriptions. Above compression patterns, denoted as $\mathbf{c}_{1}$ and $\mathbf{c}_{2}$, are preserved to guide the subsequent generation of $\mathbf{r}_{1}$ and $\mathbf{r}_{2}$.
    \item  \textbf{Reconstruction} During reconstruction, the model acquires reconstruction patterns that map user profiles back to raw trajectories, i.e., (1) How to identify components most predictive of trajectory description from user profile based on  key profile determinants identified in $\mathbf{c}_{1}$, (2) How to generate user's raw trajectory from trajectory description and candidate POIs based on $\mathbf{c}_{2}$. Above compression patterns, denoted as $\mathbf{r}_{1}$ and $\mathbf{r}_{2}$, are preserved to condition the trajectory generation process for new users.
\end{itemize}

\textbf{Mobility pattern generation. }~\label{mob_generation} As shown in Figure \ref{fig:CAMS}, in the generation phase, the model generates mobility patterns for any users  with only profile information.  To enhance the model's generalization capability, we retrieve the top k most similar template users (training users) for each new user (test user). We compare following two strategies for retrieving similarity individuals.
\begin{itemize}[leftmargin=1.5em,itemsep=0pt,parsep=0.2em,topsep=0.0em,partopsep=0.0em]
    \item \textbf{LLM-based}: Use LLM to select the top K most similar users based on semantic user profile characteristics, then directly output the ID and similarity score of each selected user.
    \item \textbf{Embedding-based}: Find similar users based on similarity scores of user profile embeddings\cite{yu2024neeko}. First, we construct a template user profile embedding matrix $\mathbf{E}_{\text{template}} \in \mathbb{R}^{m \times d}$, using the profiles of $m$ template users. Then we encode user profile of new user into an embedding $\mathbf{e}_{1\times d}$,  computing cosine similarities $\text{sim}_i$between $\mathbf{e}_{1\times d}$ and $\mathbf{E}_{\text{template}} $. Finally, we retrieve the top K users $\mathcal{T}$ with highest similarity scores.
\end{itemize}

After acquiring similar users, we sequentially perform the following steps: (1) $\mathbf{c}_{1}$ Feature Fusion: Use LLM to integrate key profile factors and high-order mobility characteristics in $\mathbf{c}_{1}$ of the similar template users. (2) Trajectory Description Generation: Using the fused features, generate trajectory descriptions by referencing $\mathbf{r}_{1}$ and $\mathbf{r}_{2}$ in compression stage. (3) $\mathbf{c}_{2}$ Feature Fusion: Use LLM to integrate both the unique movement patterns and universal movement patterns in $\mathbf{c}_{2}$ of the similar template users.

\subsection{GeoGenerator: Integrating Urban Geospatial Knowledge into Trajectory Generation}\label{geogenerator}

This section describes how to generate candidate geospatial information related to user profiles and mobility patterns. To fully leverage urban geospatial knowledge, we employ CityGPT~\cite{feng2024citygpt} as the foundation model for our agent framework, which possesses fine-grained urban spatial knowledge. Initially, Anchor Location Extractor generates critical anchor points based on user profiles, collective distributions, and geographic knowledge, which are then converted into intent-composed trajectories by incorporating mobility patterns extracted in the first stage. Subsequently, an enhanced CityGPT maps these intent-composed trajectories in real urban spaces. 
Finally, we employ further alignment CAMS with the real trajectory with the help of directly preference optimization (DPO) to further enhance the spatial continuity of generated trajectories.

\subsubsection{Anchor Location Extractor} 
The locations of homes and workplaces serve as the most important anchor points in human mobility trajectories, significantly shaped by individual user profiles and regional characteristics. To effectively identify these critical anchor locations, we propose a two-step extraction method built upon the foundation of CityGPT.

\textbf{Macro-to-micro cascaded generation.} We propose a macro-to-micro cascaded generation system with iterative reasoning-execution-reflection cycles\cite{yao2023react,du2024trajagent} to progressively refine spatial distributions. First, we transfer coordinates of all homes and workplaces into a hierarchical address representation, namely administrative area >  subdistrict >  street > POI. For regions in each hierarchy, we calculate user profile distributions and generate descriptive summaries. Then, from coarse (administrative area) to fine (street) spatial scales, the model hierarchically generates home/workplace assignments by propagating upper hierarchy outputs as contextual constraints for finer-grained reasoning. In reasoning stage, model consider descriptive summaries and geographical knowledge of child regions contained within each parent region's extent (upper hierarchy) and user profile characteristics. In execution stage,  the model select a region that best matches user profile characteristics guided by the reasoning stage. In reflection stage, the model performs periodic distribution-aware reflection.  Finally,  in POI spatial scale,  the model directly generate the precise location of home/workplace.

\textbf{Reflection with collective distribution.} We incorporate collective knowledge as feedback in reflection stage, progressively aligning generated results with distribution in real urban spaces. Upon completing execution stage of all users,  we compute spatial distribution of generated locations. Then,  in reflection stage, the model does comparative analysis against ground-truth distribution and adjusts generation strategies for subsequent iterations. Finally, in execution stage, model dynamically adjust individual output to minimize distributional divergence.

\subsubsection{Urban Structure Mapper}
To generate the remaining location points in a mobility trajectory beyond the two anchor locations (home and workplace), we introduce an Urban Structure Mapper (referred to as UrbanMapper). Given the anchor points and activity sequences, this module flexibly integrates urban spatial structure information to synthesize the remaining trajectory points.

\textbf{Enhancing CityGPT} To demonstrate the effectiveness of urban spatial knowledge, UrbanMapper leverages CityGPT injected with fine-grained urban spatial knowledge to directly generate candidate locations based on current location, user profiles, mobility patterns and intention.

To mitigate geographic hallucinations and improve spatial precision when generating specific location in real urban space, we augmented the knowledge embedded in CityGPT through fine-tuning with fine-grained urban spatial data. At a finer granularity, we posit that urban space is composed of three fundamental elements: points (POIs), lines (streets), and polygons (AOIs)\cite{dempsey2010elements}. Among these, points (POIs) constitute the most basic building blocks,  which also serve as the foundational components of trajectories. Therefore, we construct our training data based on POI-level granularity. To simulate human cognitive and exploratory processes in urban spaces, we generate navigation paths between population-weighted randomly sampled origin-destination (OD) POIs, recording all traversed POIs along the pathways, and subsequently identifying specified-category POIs within defined radius around each recorded waypoint. The radius is determined by the average jump distance between consecutive trajectory points and is correlated with the user's mobility pattern, while the category is related to user's intention at each time point. We construct a fine-tuning dataset comprising 10,000 question-answer pairs, encompassing all POIs of specified categories within certain radius around every sampled POI. 

To activate the geospatial knowledge embedded in CityGPT, we enhance user profiles with address information to infer user approximate activity ranges in real urban spaces. Furthermore, we represent the geographic elements in datasets with semantically rich addresses rather than coordinates or grid-ID. We also investigate how different address representation formats impact the model's comprehension of geographical information: (1) \textbf{Hierarchical address representation}: Use structured address hierarchies (e.g., admin→subdistrict→street→POI) to guide the model in recalling location names and attributes within specific region, reducing hallucinations and generating more realistic, specific locations. 
(2) \textbf{Human-intuitive geospatial representations}: Leverage human-intuitive geospatial representations (e.g., 100 meters from the intersection of Road B and Road C) to prompt the model to associate nearby locations and their attributes.

\textbf{Other alternative solutions. } (1) Social Graph. To model the influence of social relationships, we propose a graph-based method to provide candidate locations in real urban spaces. First, we construct a global transition graph using all historical trajectory data from training users. Let $G=(V,E)$ be the undirected graph. For each edge $e_{ij} \in E$, the weight $w_{ij}$  is computed as:
$w_{ij} = \frac{n_{ij}}{d_{ij}^\alpha + \epsilon}$,
where $n_{ij}$ means transition frequency between locations i and j, while $d_{ij}^\alpha$ means network distance (Haversine distance). A higher $w_{ij}$ indicates a greater transiting probability between two locations. Then, we identify similar users following the methodology proposed in section \ref{mob_generation}. Candidate locations for user to visit next are determined by the most likely next locations visited by similar users, which are reasoned on the graph. (2) Map Tools. To evaluate the effectiveness of commercial geospatial APIs, we construct a mapping between intentions and location categories and get candidate locations for user to visit next through map queries. The radius and the mapping relationship are fixed regardless of variation of mobility patterns.

\subsection{TrajEnhancer: Enhanced Trajectory Generation with Preference Alignment}\label{trajenhancer}

TrajEnhancer performs integrated reasoning by synthesizing the  urban spatial knowledge generated in section~\ref{geogenerator} and mobility patterns extracted in section~\ref{mob_generation}. It first generates daily activity plans for target users based on their profiles and mobility patterns, which consists of intentions and temporal constraints. Subsequently, it synthesizes realistic movement trajectories by holistically considering user profiles, mobility patterns, activity plans, anchors points, and urban geospatial knowledge.

To enhance the spatiotemporal continuity of the model-generated trajectories, we apply iterated DPO training to further enrich the model's urban geographic knowledge and enhance it's ability to identify mobility patterns. We construct the training dataset using the corpus output by CAMS and corresponding individuals' real trajectories. We execute multiple cycles of training → deployment → testing → data collection → retraining. Through these progressive multi-phase training iterations, we aim to progressively enhance CityGPT's comprehension of the relationship between user mobility patterns and the spatiotemporal attributes of trajectory points in real urban spaces, thereby fully activating its spatiotemporal reasoning ability for mobility simulation.

\section{Experiments}

\subsection{Experimental Setup}

\paragraph{Datasets}
We carry out experiment using two real-world mobility datasets, ChinaMobile and Tencent. The basic information of the datasets is shown in Table \ref{tab:dataset-info}. To test CAMS's performance on public datasets, we employ open street map's road network data and AOI data along with global POI data from Foursquare to jointly represent urban spaces. This does not compromise the overall experimental results. This confirms the transferability of CAMS across different datasets, and can achieve reasonably good performance even on smaller, lower-quality datasets.

\begin{table}[t]
  \centering
  \small
  \caption{Basic information of the trajectory datasets}
  \label{tab:dataset-info}
  \resizebox{0.9\linewidth}{!}{%
  \begin{tabular}{lcc}
    \toprule
    \textbf{Datasets} & \textbf{Tencent} & \textbf{ChinaMobile} \\ \midrule
    Duration & October 1, 2019 -- December 31, 2019 & July 1, 2017 -- August 31, 2017 \\ 
    City & Beijing & Beijing \\ 
    \#Users & 100,000 & 1,246 \\ 
    \#Trajectory Points & 297,363,263 & 4,163,651 \\ \bottomrule
  \end{tabular}%
  \vspace{-10pt}
  }
\end{table}
\textbf{Metrics.} Following previous work~\cite{feng2020learning, shao2024beyond}, we evaluate the quality of generated mobility data from three dimensions, including statistical evaluation, aggregation evaluation and semantics evaluation. We also use Toponym Valid Ratio (TVR) to measure geographic knowledge hallucination, and Composite Mean Reciprocal Rank (CMRR) to measure overall performance across all metrics. 
\begin{itemize}[leftmargin=1.5em,itemsep=0pt,parsep=0.2em,topsep=0.0em,partopsep=0.0em]
    \item \textbf{Individual evaluation.} We calculate Jensen–Shannon Divergence (JSDs) on the following metrics of per user: Distance, Radius, Step Interval (SI), Step Distance (SD) and Spatial-temporal Visits Distribution (STVD).
    
    \item \textbf{Collective evaluation.} We evaluate the quality of all generated data from a collective perspective, calculating JSDs on following metric of all users: Frequently visited locations (FVLoc), which is defined as the overall distributions of  top 40 most frequently visited locations across all users.
    
    \item \textbf{Semantics evaluation.} To evaluate the plausibility of generated mobility data, we map venue categories to user intents (e.g., Food → dining) and compute JSDs on following category-related metrics at both individual and collective levels: Daily Activity Routine Distribution (DARD) and Activity Probability (ActProb).

    \item \textbf{Hallucination evaluation.} We define Toponym Valid Ratio (TVR), which is the ratio of valid generated toponyms to total generated toponyms, to assess the degree of hallucination in the model's candidate geospatial knowledge generation.

    \item \textbf{Comprehensive evaluation.} To holistically assess model performance, we propose the Composite Mean Reciprocal Rank (CMRR) metric, computed through a two-stage process: (1) calculating the reciprocal rank of each metric relative to all comparable models, then (2) computing the arithmetic mean of these reciprocal ranks across all metrics. 
\end{itemize}
\textbf{Methods.} We compared our model against several state-of-the-art approaches: three deep-learning-based models including ActSTD\cite{yuan2022activity}, DSTPP\cite{yuan2023spatio} and MoveSim\cite{feng2020learning}, two LLM-based models including CoPB\cite{shao2024beyond} and LLMob\cite{wang2024simulating}, and a classic mechanistic model TimeGeo\cite{jiang2016timegeo}.

\subsection{Main Results}

\subsubsection{Mobility generation in real urban space}
We want to validate the model's ability to generate geospatially accurate trajectories in real-world urban space using only minimal user profile data. To ensure fair comparison, for LLM-based models, we employ llama3.1-8b as LLM core, while removing all specific location names (except anchor points) and manually extracted user-specific trajectory features from prompts; for deep-learning-based methods, we apply uniform time-interval interpolation to sparse trajectory points, and reduce training set size to 3× test set to simulate limited urban context.

The experimental results demonstrate that CAMS exhibits more pronounced advantages in trajectory generation phase, achieving superior performance on 11 out of 16 metrics with the highest CMRR score. It performs exceptionally well on metrics related to spatial distributions (Distance, Radius, and SD), which can be attributed to its effective utilization of built-in urban spatial knowledge. Furthermore, CAMS maintains its leading performance on individual mobility pattern metrics (DARD), indicating its strong generalization capability to infer new users' mobility patterns accurately based solely on profile information. The model also shows competitive results on collective distribution metrics (FVLoc, ActProb), suggesting its effective consideration of how trajectories with different user profiles distribute in real urban spaces. Compared to mobility recovery results , CAMS maintains consistent performance while other models experience significant performance degradation under limited input information, further highlighting its remarkable transfer learning capability in zero-shot scenarios.

To demonstrate that CityGPT effectively captures the relationships between user profiles, mobility patterns, and trajectories in real urban spaces, we visualized the anchor points and single-day trajectory point distributions for different user profiles generated by the model in Figure \ref{fig:spatial-patterns}. We conduct a comparative analysis of trajectory points distribution characteristics across different user profiles generated by the model (see Appendix~\ref{user mobility patterns} for details).

\begin{table}[t]
  \centering
  \small
  \caption{Performance comparison of mobility simulation methods across datasets. Best and second-best results are highlighted in \textbf{bold} and \underline{underline}, respectively.}
  \label{tab:generate-results}
  \resizebox{\linewidth}{!}{%
  \begin{tabular}{ll|cccccccc|c}
    \toprule
    \multirow{2}{*}{Dataset} & \multirow{2}{*}{Model} & \multicolumn{8}{c}{Generation Metrics} & \multirow{2}{*}{CMRR$\uparrow$} \\ 
    \cmidrule{3-10}
    & & FVLoc$\downarrow$ & ActProb$\downarrow$ & Distance$\downarrow$ & Radius$\downarrow$ & SI$\downarrow$ & SD$\downarrow$ & DARD$\downarrow$ & STVD$\downarrow$ & \\ 
    \midrule
    \multirow{7}{*}{Tencent} 
    & TimeGeo & 0.3671 & 0.5915 & 0.3053 & 0.3267 & 0.2312 & 0.3414 & 0.6972 & 0.6899 & 0.1798 \\ 
    & ActSTD  & 0.3518 & 0.6972 & 0.2815 & \underline{0.1007} & 0.6880 & 0.0907 & 0.6946 & 0.6371 & 0.2574 \\
    & DSTPP & \underline{0.3115} & 0.5236 & \textbf{0.1599} & 0.1589 & 0.6880 & 0.0308 & 0.6863 & 0.6742 & \underline{0.4458} \\ 
    & MoveSim  & 0.3650 & \textbf{0.2647} & 0.3876 & 0.3192 & \underline{0.2309} & 0.1716 & 0.4946 & \textbf{0.5271} & 0.4354 \\ 
    & COPB  & \textbf{0.2744} & 0.3907 & \underline{0.1997} & 0.2394 & 0.4131 & 0.1915 & \underline{0.4089} & \underline{0.5660} & 0.4146 \\
    & LLMob & 0.5044 & 0.3723 & 0.5266 & 0.1723 & \textbf{0.1618} & \underline{0.0752} & 0.4400 & 0.6639 & 0.3420 \\
    & CAMS & 0.3213 & \underline{0.2769} & 0.2596 & \textbf{0.0517} & 0.3362 & \textbf{0.0457} & \textbf{0.3832} & 0.6540 & \textbf{0.5208} \\ 
    \midrule
    
    \multirow{7}{*}{ChinaMobile} 
    & TimeGeo & 0.3825 & 0.6793 & 0.4009 & 0.5466 & 0.5139 & 0.3065 & 0.5762 & 0.6907 & 0.1940 \\ 
    & ActSTD  & 0.4025 & 0.6900 & 0.4577 & 0.3400 & 0.5749 & \textbf{0.0452} & 0.6891 & 0.6947 & 0.2673 \\
    & DSTPP & 0.3895 & 0.6794 & 0.3991 & 0.2553 & 0.5320 & 0.0989 & 0.6774 & 0.6613 & 0.2229 \\
    & MoveSim  & 0.3776 & 0.2413 & 0.5117 & 0.3810 & 0.3720 & 0.1063 & 0.5264 & \textbf{0.5775} & 0.3479 \\  
    & COPB & \textbf{0.2724} & \underline{0.2097} & \underline{0.2690} & 0.1948 & 0.4794 & 0.2282 & 0.4997 & \underline{0.5817} & \underline{0.4479} \\
    & LLMob & 0.5130 & 0.4198 & 0.5387 & \underline{0.1690} & \textbf{0.1207} & 0.0839 & \underline{0.3976} & 0.6585 & 0.4003 \\
    & CAMS & \underline{0.2994} & \textbf{0.1244} & \textbf{0.1867} & \textbf{0.0507} & \underline{0.2524} & \underline{0.0473} & \textbf{0.3544} & 0.6776 & \textbf{0.7125} \\ 
    \bottomrule
  \end{tabular}%
  }
  \vspace{-10pt}
\end{table}

\subsubsection{Semantic-based mobility recovery}
First we want to validate the effectiveness of MobExtractor. During compression, CAMS extracts the relationships between original trajectories and user profiles. In the reconstruction phase, these user profiles are utilized to reconstruct the original trajectories under the guidance of mobility patterns extracted from relationships. Lower Jensen-Shannon Divergence(JSD) scores between recovered trajectories and original trajectories indicates better performance of recovery stage of MobExtractor. The results of evaluating performance are detailed in Table \ref{tab:recovery-results}. The results demonstrate that despite using only user profiles as external input, CAMS achieves superior performance on 7 of 16 metrics in both data sets, with particularly outstanding advantages in metrics evaluating individual mobility capability (Radius) and behavioral habits (DARD). CAMS also exhibits commendable performance in terms of spatial continuity within trajectories (Distance and SD). Comparative analysis revealed that LLMob performs best on features related to collective distribution and semantics (FVLoc) as well as individual routine patterns (SD), while MoveSim shows better results on metrics measuring collective distribution (FVLoc and ActProb). CAMS does not demonstrate significant advantages in these particular aspects. We attribute LLMob's strengths to its model architecture's emphasis on its explicit incorporation of personal movement characteristics as external input, whereas MoveSim's advantages stem from its inherent data-driven approach that better fits overall distributions. However, both methods underperform significantly compared to CAMS on metrics evaluating individual mobility behaviors in real urban spaces (Distance, Radius). This superior performance of CAMS can be attributed to its comprehensive consideration of the alignment between urban geographical knowledge and user mobility patterns. For instance, when analyzing a low-income migrant worker with more constrained and fixed mobility patterns, the model preferentially considers workplace locations (areas with concentrated large factories), residential areas (neighborhoods with lower living costs), and nearby dining and entertainment venues during the generation process.

\begin{table}[t]
  \centering
  \small
  \caption{Performance comparison of trajectory recovery methods across datasets. Best and second-best results are highlighted in \textbf{bold} and \underline{underline}, respectively.}
  \label{tab:recovery-results}
  \resizebox{\linewidth}{!}{%
  \begin{tabular}{ll|cccccccc|c}
    \toprule
    \multirow{2}{*}{Dataset} & \multirow{2}{*}{Model} & \multicolumn{8}{c}{Recovery Metrics} & \multirow{2}{*}{CMRR$\uparrow$} \\ 
    \cmidrule{3-10}
    & & FVLoc$\downarrow$ & ActProb$\downarrow$ & Distance$\downarrow$ & Radius$\downarrow$ & SI$\downarrow$ & SD$\downarrow$ & DARD$\downarrow$ & STVD$\downarrow$ & \\ 
    \midrule
    \multirow{7}{*}{Tencent} 
    & TimeGeo & 0.3555 & 0.6673 & 0.1506 & 0.2749 & 0.2010 & 0.3362 & 0.6895 & 0.6921 & 0.1890 \\
    & ActSTD & 0.2007 & 0.6737 & \textbf{0.0841} & \underline{0.0581} & 0.6880 & \underline{0.0606} & 0.6910 & 0.6746 & 0.3494 \\
    & DSTPP & 0.1975 & 0.3582 & 0.1346 & 0.2611 & 0.3960 & 0.0775 & 0.4824 & \underline{0.4233} & 0.2792 \\
    & MoveSim & \underline{0.1966} & \underline{0.1647} & 0.3176 & 0.3189 & \underline{0.2009} & 0.1315 & 0.4456 & \textbf{0.4141} & 0.4045 \\ 
    & COPB & 0.1981 & 0.3592 & \underline{0.1322} & 0.2270 & 0.3858 & 0.1885 & \underline{0.3930} & 0.5474 & 0.3000 \\
    & LLMob & \textbf{0.1213} & \textbf{0.0891} & 0.1822 & 0.1330 & \textbf{0.1618} & 0.0630 & 0.3961 & 0.5923 & \textbf{0.5563} \\
    & CAMS & 0.3421 & 0.2444 & 0.2087 & \textbf{0.0340} & 0.2945 & \textbf{0.0443} & \textbf{0.2681} & 0.6284 & \underline{0.5146} \\ 
    \midrule
    
    \multirow{7}{*}{ChinaMobile} 
    & TimeGeo & 0.3766 & 0.6862 & 0.3779 & 0.2908 & 0.4517 & 0.3057 & 0.6874 & 0.6845 & 0.1619 \\ 
    & ActSTD & 0.2985 & 0.6862 & 0.4336 & 0.2816 & 0.5321 & \textbf{0.0324} & 0.6905 & 0.6907 & 0.2631 \\
    & DSTPP & 0.2136 & 0.3722 & 0.1570 & 0.2705 & \underline{0.2941} & 0.1537 & 0.5692 & \textbf{0.4350} & 0.3708 \\
    & MoveSim & \underline{0.1776} & \textbf{0.1496} & 0.3300 & 0.2619 & 0.4514 & 0.0950 & 0.4346 & \underline{0.4617} & 0.4104 \\ 
    & COPB & 0.2677 & 0.2287 & \underline{0.1415} & 0.1948 & 0.4735 & 0.1945 & 0.4790 & 0.5474 & 0.2917 \\
    & LLMob & \textbf{0.1560} & \underline{0.1590} & 0.1654 & \underline{0.0763} & \textbf{0.1207} & 0.0718 & \underline{0.3795} & 0.6560 & \underline{0.5417} \\
    & CAMS & 0.3055 & 0.3664 & \textbf{0.1339} & \textbf{0.0717} & 0.3464 & \underline{0.0484} & \textbf{0.3248} & 0.6601 & \textbf{0.5563} \\ 
    \bottomrule
  \end{tabular}%
  }
  \vspace{-10pt}
\end{table}

\begin{table}[htbp]
  \centering
  \small
  \vspace{-10pt}
  \caption{Performance comparison of different LLMs within the CAMS framework. Best and second-best results are highlighted in \textbf{bold} and \underline{underline}, respectively.}
\label{tab:model-results}
  \resizebox{\linewidth}{!}{%
  \begin{tabular}{ll|ccccccccc}
    \toprule
    \multirow{2}{*}{Dataset} & \multirow{2}{*}{Base Model} & \multicolumn{9}{c}{Generation Metrics} \\ 
    \cmidrule{3-11}
    & & FVLoc$\downarrow$ & ActProb$\downarrow$ & Distance$\downarrow$ & Radius$\downarrow$ & SI$\downarrow$ & SD$\downarrow$ & DARD$\downarrow$ & STVD$\downarrow$ & TVR$\uparrow$ \\ 
    \midrule
    \multirow{8}{*}{Tencent} 
    & LLaMA3.1-8B & 0.4315 & 0.4649 & 0.3109 & 0.0920 & 0.2751 & 0.0883 & 0.3810 & 0.6672 & \underline{0.9570} \\
    & LLaMA3-70B & 0.4342 & 0.3102 & \underline{0.2985} & \underline{0.0529} & 0.2053& \underline{0.0460} & \textbf{0.2896} & 0.6573& 0.9560\\
    & Qwen2-72B & 0.4119& 0.3421 & 0.4384 & 0.1028 & 0.2128 & 0.1072 & 0.3203 & 0.6601 & 0.8420 \\
    & Qwen3-235B & 0.4089 & \underline{0.2738}& 0.3582 & 0.1015 & 0.2899 & 0.1725 & 0.3569 & \textbf{0.6416}& 0.8247 \\
    & GPT-4o-mini & 0.4119 & \textbf{0.2499} & 0.3753 & 0.1143 & \underline{0.1874}& 0.1218 & \underline{0.3046} & 0.6624 & 0.8672 \\
    & Gemma3-27B & \underline{0.3994}& 0.3903 & 0.4160 & 0.0695 & \textbf{0.1717}& 0.0506 & 0.3265 & 0.6570 & 0.8252 \\
    & Mistral7Bv3 & 0.4089 & 0.3547 & 0.3064 & 0.0811 & 0.3101 & 0.0980 & 0.3574 & 0.6677 & 0.8854 \\
    & CityGPT & \textbf{0.3213} & 0.2769& \textbf{0.2596} & \textbf{0.0517} & 0.3362 & \textbf{0.0457} & 0.3832 & \underline{0.6540}& \textbf{1.0000} \\
    \midrule
    
    \multirow{8}{*}{ChinaMobile} 
    & LLaMA3.1-8B & \underline{0.3992} & 0.4434 & 0.4045 & 0.0640 & 0.2521 & 0.0731 & 0.3987 & 0.6788 & \underline{0.9684} \\
    & LLaMA3-70B & 0.4059 & 0.4120& \underline{0.3452} & 0.0638 & \underline{0.2031}& 0.0517 & \underline{0.3372}& 0.6792 & 0.9650 \\
    & Qwen2-72B & 0.4027 & 0.4414 & 0.3770 & \underline{0.0618} & 0.2164& \underline{0.0479} & 0.3525& 0.6780 & 0.8535 \\
    & Qwen3-235B & 0.3992 & 0.5140 & 0.4254 & 0.0937 & 0.2976 & 0.0949 & 0.3791 & \underline{0.6586}& 0.8228 \\
    & GPT-4o-mini & 0.4027 & 0.4493 & 0.4894 & 0.0714 & 0.2802 & 0.0630 & 0.3689 & 0.6644& 0.7836 \\
    & Gemma3-27B & 0.3994 & \underline{0.3903}& 0.4160 & 0.0695 & \textbf{0.1717}& 0.0506 & \textbf{0.3265}& \textbf{0.6570}& 0.8252 \\
    & Mistral7Bv3 & 0.4059 & 0.4344 & 0.3806 & 0.0913 & 0.2035 & 0.0822 & 0.3608 & 0.6821 & 0.9007 \\
    & CityGPT & \textbf{0.2994} & \textbf{0.1244} & \textbf{0.1867} & \textbf{0.0507} & 0.2524 & \textbf{0.0473} & 0.3544 & 0.6776& \textbf{1.0000} \\
    \bottomrule
  \end{tabular}%
  }
\end{table}

\subsection{Ablation Studies}
In this section, we perform analysis on varying model designs to further demonstrate the rationality and effectiveness of the model design. We also compared the task performance of CityGPT with other open-source/closed-source LLMs, further demonstrating the effectiveness of CityGPT in providing user-relevant urban geospatial knowledge.

\begin{figure*}[!t]
  \centering
  \begin{minipage}[t]{0.45\linewidth}
    \centering
    \includegraphics[width=\linewidth, height=0.6\linewidth]{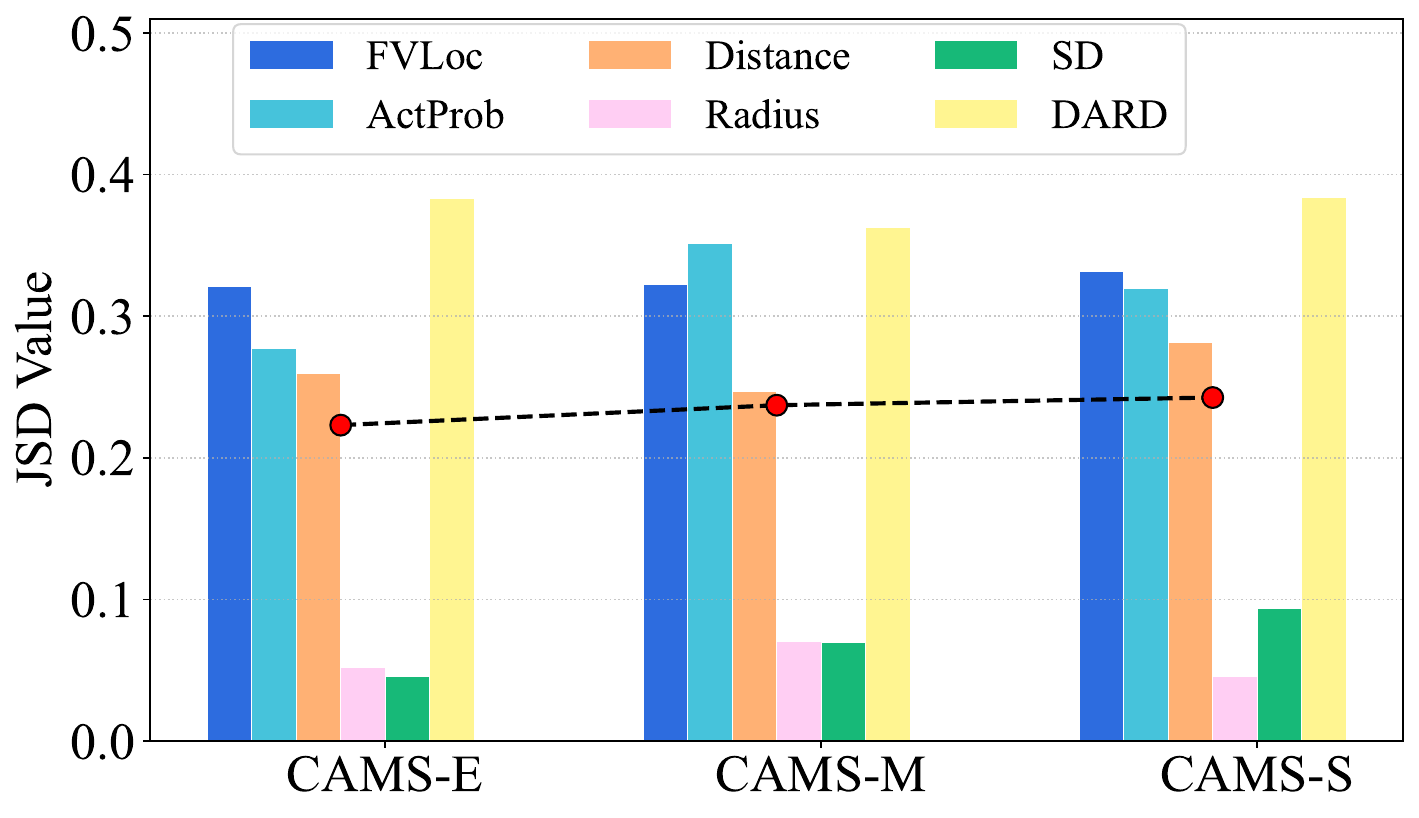}
    \caption{Methodological comparisons in UrbanMapper using Tencent dataset}
    \vspace{-10pt}
    \label{fig:tencent-metric-comparation}
  \end{minipage}
  \hfill
  \begin{minipage}[t]{0.45\linewidth}
    \centering
    \includegraphics[width=\linewidth, height=0.6\linewidth]
    {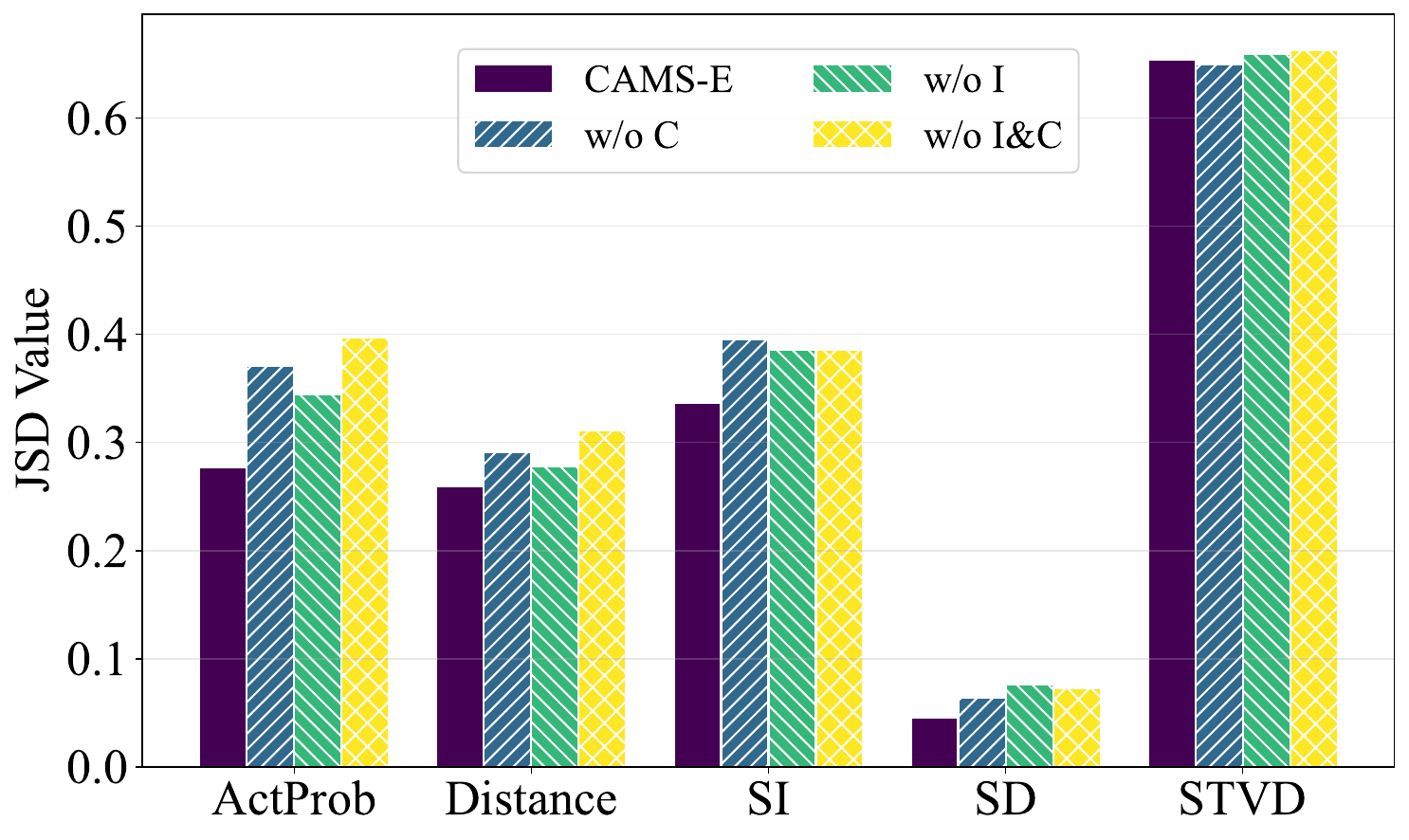}
    \caption{Ablation study on model designs(generation phase).}
    \vspace{-10pt}
    \label{fig:ablation-results}
  \end{minipage}
\end{figure*}

\textbf{Impact of reflection in Anchor Location Extractor.} As we introduce in section \ref{geogenerator}, we incorporate collective knowledge as feedback in reflection stage of Anchor Location Extractor. By analysing recovery results in Table \ref{tab:ablation-results} and generation results in Figure  \ref{fig:ablation-results}, we observe that the reflective version consistently outperforms its non-reflective counterpart(w/o C) across all metrics. This improvement confirms that by integrating collective knowledge, the model can more accurately infer the relationship between user profiles and real-world urban spatial patterns, consequently generating trajectories that better align with actual urban mobility distributions.

\begin{figure*}[t]
  \centering
  \begin{subfigure}[t]{0.3\textwidth}
    \centering
    \includegraphics[width=\linewidth, height=\linewidth]{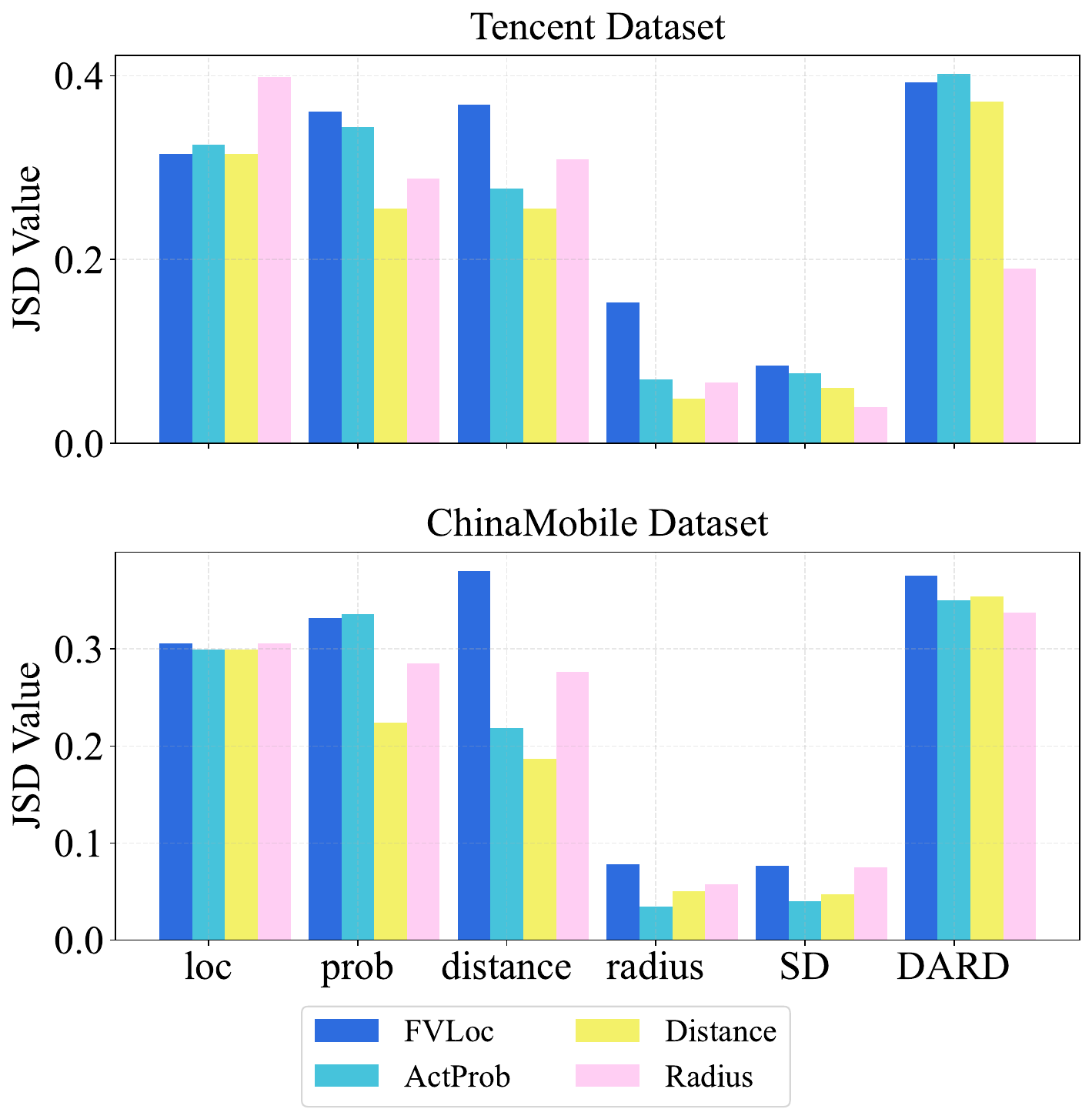}
    \caption{Variations in metrics}
    \label{fig:trajenhancer-allmetrics}
  \end{subfigure}%
  \hfill
  \begin{subfigure}[t]{0.3\textwidth}
    \centering
    \raisebox{0.1\linewidth}{  %
      \includegraphics[width=\linewidth, height=0.8\linewidth]{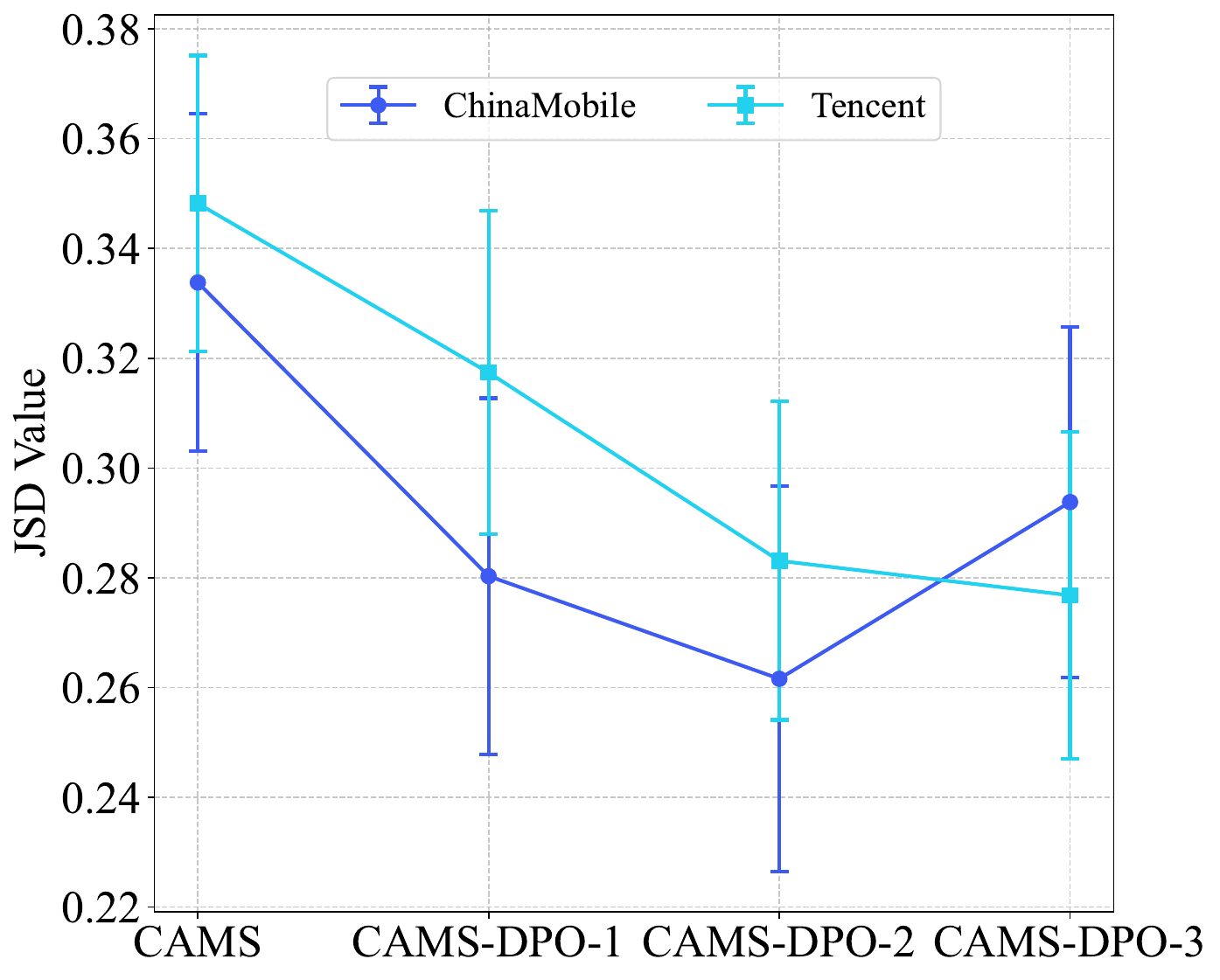}
    }
    \caption{Overall performance}
    \label{fig:trajenhancer-overall}
  \end{subfigure}%
  \hfill
  \begin{subfigure}[t]{0.3\textwidth}
    \centering
    \includegraphics[width=\linewidth, height=\linewidth]{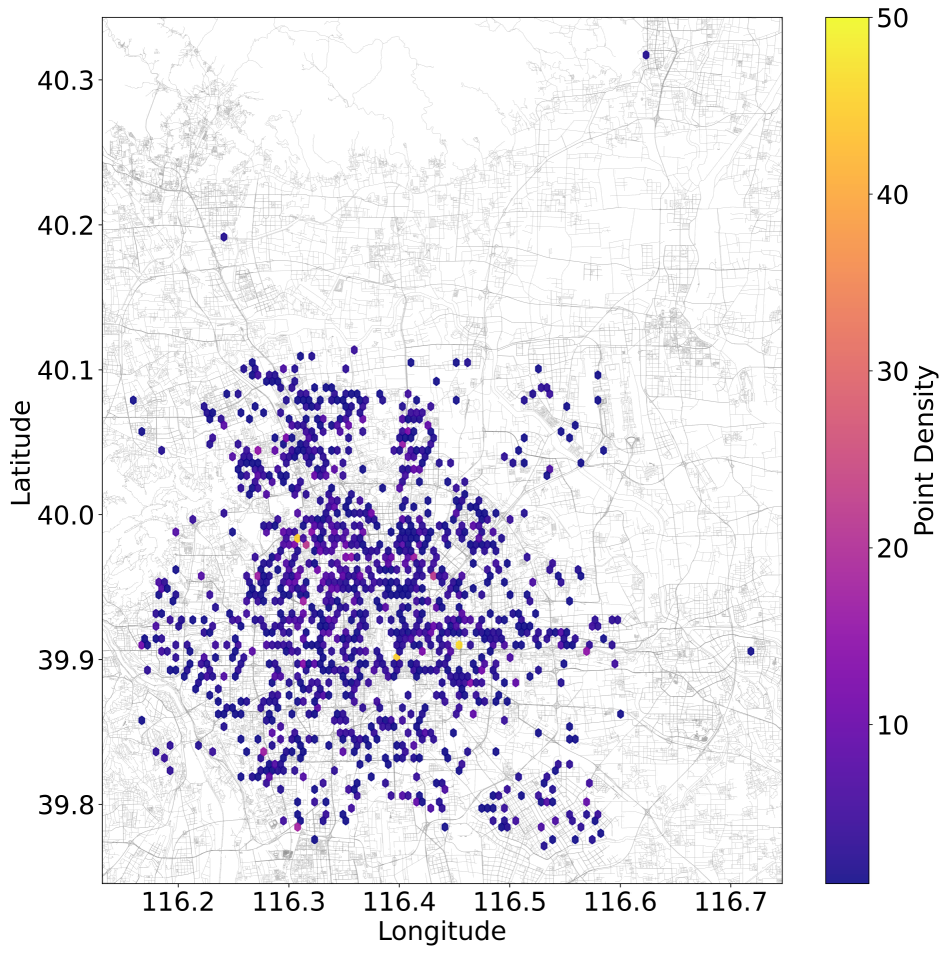}
    \caption{Trajectory visualization} %
    \label{fig:trajenhancer-visual}
  \end{subfigure}
  \caption{DPO result analysis of TrajEnhancer.}
  \vspace{-10pt}
  \label{fig:trajenhancer-results}
\end{figure*}

\textbf{Impact of TrajEnhancer.} We evaluate overall performance of trajectory enhancement module in Table \ref{tab:ablation-results}. As visually confirmed in Figure \ref{fig:trajenhancer-overall}, there is an overall reduction in JSDs across successive DPO iterations, indicating that TrajEnhancer progressively enhances the spatiotemporal continuity of generated trajectories to approximate real-world mobility patterns. Variations of each metric are visualized in Figure \ref{fig:trajenhancer-allmetrics}.

\textbf{Comparison of different methodologies in UrbanMapper.} By comparing results of using CityGPT-enhanced(CAMS-E), map tools(CAMS-M) and social networks(CAMS-S) in Figure \ref{fig:tencent-metric-comparation}, we find that CAMS-E outperforms other methods with visibly lower JSDs. This suggests that implicitly incorporating geographic knowledge in trajectory generation tasks is reasonable, and CityGPT offers greater advantages over traditional GIS tools and social relationships.

\textbf{Performance comparison between enhanced CityGPT and other LLMs.} We test the performance of multiple open-source and closed-source LLMs in experimental scenarios. The results in Table \ref{tab:model-results} demonstrate that CityGPT, based on the Llama3.1-8B pre-trained model, can provide more authentic and fine-grained urban geospatial knowledge compared to other larger-parameter models. Additionally, CityGPT achieves the highest CMRR, indicating its superior ability to capture the connections between user profiles, mobility patterns and geospatial knowledge.

\vspace{-10pt}

\section{Related Work}
\vspace{-10pt}
\label{gen_inst}

\textbf{Mobility simulation.} 
On the basis of macroscopic statistical laws~\cite{Brockmann2006scaling, roth2011structure,liang2012scaling}, 
researchers proposed a series of  mobility simulation models to depict individual behavior mechanism~\cite{pappalardo2018data, wang2019extended, pappalardo2018data, jiang2016timegeo}. While these mechanism models are concise but fail to capture complex human mobility patterns and model the impact of urban structure. With the rapid development of deep learning, different model structures were designed to model the complex dynamics of mobility behaviors~\cite{feng2020learning, long2023practical, liu2024act2loc}. 
However, these deep learning methods face challenges of data sparsity, poor transferability and low explainability. 

\textbf{LLM for geospatial tasks.} 
Since LLMs are geospatially knowledgeable~\cite{ChatGPT, bhandari2023large, touvron2023llama}, researchers pay attention to leverage LLM in geography and urban science field by solving domain-specific tasks like geospatial understanding tasks~\cite{brown2020language, mai2023opportunities, roberts2023gpt4geo, feng2024citygpt} and geospatial prediction tasks~\cite{wang2023would, beneduce2024large, feng2024agentmove, gong2024mobility}. LLMs can achieve good results in global-scale or national-scale tasks with simple prompt engineering~\cite{manvi2023geollm, manvi2024large} or a trained linear layer~\cite{gurnee2023language}. However, when breaks down to city scale, well-designed agentic frameworks and fine-tuning approaches are required to enable LLMs to acquire urban structural knowledge~\cite{feng2024citygpt, balsebre2024lamp} and enhance task-specific performance via geospatial knowledge alignment.

\textbf{LLM for mobility simulation.} With the successful application of LLM in geospatial tasks, researchers are exploring the potential of applying LLMs to human mobility simulation~\cite{jiawei2024large, shao2024beyond, wang2024simulating, li2024more, jiang2024towards, yan2024opencity}. They extract individual knowledge from the user profile and historical trajectories, then synthesize simulated data~\cite{wang2024simulating}, map simulated data to real urban spaces using mechanistic models~\cite{shao2024beyond,li2024more}, or generate real-world trajectories based on the given urban spatial information~\cite{jiawei2024large}. They perform well in few-shot scenarios and exhibit good transferability. However, they insufficiently model real urban structures, and fail to effectively capture collective mobility patterns.

\section{Conclusion}
In this paper, we propose \ourmodel, an innovative agentic framework for generating realistic urban mobility trajectories by building upon CityGPT. \ourmodel~utilizes urban foundation model's inherent geospatial knowledge while incorporating advanced commonsense reasoning techniques to capture underlying movement patterns. We design MobExtractor that synthesize mobility patterns via semantic comprehension, encoding and feature fusion mechanisms. We develop GeoGenerator to enhance CityGPT's urban spatial intelligence, and TrajEnhancer to improve spatiotemporal continuity in trajectory generation. Extensive experiments on real-world trajectory datasets demonstrate the framework's capability to directly generate realistic trajectories from new user profiles.

\bibliographystyle{plainnat}  %
\bibliography{references}

\newpage
\appendix
\section{Appendix}
\subsection{Data preprocessing}
\paragraph{Trajectory data preprocessing}
 Among the 100,000 users in the Tencent dataset, 44,313 users have both complete user profiles and accurate coordinates for their home and workplace. We conduct Collective Knowledge Extractor experiment using this subset of data. For each mobility dataset, we select the top 150 users with the highest average daily trajectory points for Geographic Knowledge Extraction,  among which  we randomly assign 100 users for trajectory recovery and the remaining 50 users for trajectory generation. We utilize the top 1,500 users with relatively higher average daily trajectory points for Individual Knowledge Extractor.

We aligned the private dataset with public datasets by replacing the private dataset's urban elements (including AOIs, POIs and roads) with relevant elements in OSM and Foursquare. Notably, while the aligned mobility dataset becomes sparser, this does not compromise the overall experimental results. This confirms the transferability of CAMS accross different datasets, and can achieve reasonably good performance even on smaller, lower-quality datasets.

\paragraph{DPO training data construction} We construct the training dataset using the corpus output by CAMS. We employ Qwen2-72B, a large language model primarily trained on Chinese corpora, to extract mobility patterns while leveraging CityGPT for urban geographical knowledge. Our experimental setup involves 300 training users and 1200 test users, with all textual outputs from CAMS being collected for analysis. Using Qwen2.5-72B, we evaluate the quality of mobility pattern extraction from this corpus by assigning quality scores on a 0-10 scale, where we select textual outputs scoring above 5 as negative samples while using the corresponding individuals' real trajectories as positive samples for our training data construction.

\subsection{Baselines}
\begin{itemize}[leftmargin=1.5em,itemsep=0pt,parsep=0.2em,topsep=0.0em,partopsep=0.0em]
    \item \textbf{Deep-learning-based models} (1) ActSTD\cite{yuan2022activity}: It adopts a GAIL framework, combining continuous spatio-temporal dynamics modeling with generative adversarial training to generate mobility data. (2) DSTPP\cite{yuan2023spatio}: It employs a diffusion model to learn the joint spatio-temporal distribution, incorporating a spatio-temporal co-attention module for modeling Spatio-Temporal Point Processes (STPP).  (3) MoveSim\cite{feng2020learning}: It adopts a generative-adversarial framework in which the generator utilizes a self-attention-based sequence modeling network to capture temporal transitions and the discriminator distinguishes synthetic trajectories by incorporating key mobility patterns.
    \item \textbf{LLM-based models} (1) CoPB\cite{shao2024beyond}: It leverages LLM to infer mobility-related habits and motivations from user profiles, then apply mechanistic models to map these patterns to real urban spaces. (2)LLMob\cite{wang2024simulating} : It derives mobility patterns from predefined trajectory features using LLM, then selects positive/negative training samples from historical mobility data for adversarial learning.
    \item \textbf{Mechanistic model} (1) TimeGeo\cite{jiang2016timegeo}: It uses statistical methods to model temporal patterns while leveraging the r-EPR mechanism to model spatial patterns of user mobility data.
\end{itemize}

\subsection{Ablation studies}

\begin{table}[htbp]
  \centering
  \small
  \caption{Performance comparison of different LLMs in trajectory recovery within the CAMS framework. Best and second-best results are highlighted in \textbf{bold} and \underline{underline}, respectively.}
  \label{tab:model-results-recovery}
  \resizebox{\linewidth}{!}{%
  \begin{tabular}{ll|ccccccccc}
    \toprule
    \multirow{2}{*}{Dataset} & \multirow{2}{*}{Base Model} & \multicolumn{9}{c}{Recovery } \\ 
    \cmidrule{3-11}
    & & FVLoc$\downarrow$ & ActProb$\downarrow$ & Distance$\downarrow$ & Radius$\downarrow$ & SI$\downarrow$ & SD$\downarrow$ & DARD$\downarrow$ & STVD$\downarrow$ & TVR$\uparrow$ \\ 
    \midrule
    \multirow{8}{*}{Tencent} 
    & LLaMA3.1-8B & \underline{0.3468} & \underline{0.2776} & \underline{0.1913} & \underline{0.0362} & 0.3070 & 0.0794 & 0.3300 & 0.6774 & \underline{0.9697} \\
    & LLaMA3-70B & 0.4468 & 0.3180 & \textbf{0.1770} & 0.0529 & 0.1697 & \textbf{0.0383} & 0.2684 & 0.6310 & 0.9541 \\
    & Qwen2-72B & 0.4421 & 0.3567 & 0.2177 & 0.0785 & 0.1682& 0.0881 & 0.2848 & \textbf{0.6091} & 0.8224 \\
    & Qwen3-235B & 0.4468 & 0.3152 & 0.1646 & 0.0687 & 0.2819 & 0.0895 & 0.3255 & 0.6178 & 0.9040 \\
    & GPT-4o-mini & 0.4468 & 0.3682 & 0.2737 & 0.1022 & \underline{0.1650}& 0.1038 & \textbf{0.2628} & \underline{0.6170} & 0.7844 \\
    & Gemma3-27B & 0.4468 & 0.3134 & 0.3243 & 0.0887 & \textbf{0.1637}& 0.0964 & 0.2789 & 0.5967 & 0.8477 \\
    & Mistral7Bv3 & 0.4468 & 0.2799 & 0.2273 & 0.0639 & 0.2030 & 0.0865 & 0.2886 & 0.6265 & 0.9035 \\
    & CityGPT-E & \textbf{0.3421} & \textbf{0.2444} & 0.2087 & \textbf{0.0340} & 0.2945 & \underline{0.0443} & \underline{0.2681} & 0.6284 & \textbf{1.0000} \\
    \midrule
    
    \multirow{8}{*}{ChinaMobile} 
    & LLaMA3.1-8B & \underline{0.3176} & 0.3964 & 0.2930 & 0.0987 & 0.2400 & 0.0612 & 0.3760 & 0.6769 & \underline{0.9815} \\
    & LLaMA3-70B & 0.4118 & \underline{0.3826} & 0.2746& \underline{0.0753} & \underline{0.1953}& \underline{0.0501} & 0.3333 & 0.6615 & 0.9690 \\
    & Qwen2-72B & 0.4118 & 0.4025 & 0.3616 & 0.1050 & 0.2190 & 0.0864 & 0.3373 & 0.6535& 0.8306 \\
    & Qwen3-235B & 0.4118 & 0.4242 & 0.2852 & 0.0770 & 0.3615 & 0.0564 & 0.4550 & 0.6595 & 0.9310 \\
    & GPT-4o-mini & 0.4118 & 0.4666 & 0.3737 & 0.1285 & 0.2175& 0.0803 & \underline{0.3276} & \textbf{0.6468} & 0.8561 \\
    & Gemma3-27B & 0.4118 & 0.4752 & 0.3390 & 0.0871 & \textbf{0.1684}& 0.0883 & 0.3346 & \underline{0.6480}& 0.8436 \\
    & Mistral7Bv3 & 0.4118 & 0.5327 & \underline{0.2345}& 0.0876 & 0.2598 & 0.0665 & 0.3582 & 0.6488 & 0.9164 \\
    & CityGPT-E & \textbf{0.3055} & \textbf{0.3664} & \textbf{0.1339} & \textbf{0.0717} & 0.3464 & \textbf{0.0484} & \textbf{0.3248} & 0.6601 & \textbf{1.0000} \\
    \bottomrule
  \end{tabular}%
  }
\end{table}

\begin{table}[htbp]
  \centering
  \small
  \caption{Performance comparison of different methodology variants in UrbanMapper. Best and second-best results are highlighted in \textbf{bold} and \underline{underline}, respectively.}
  \label{tab:ablation-results}
  \resizebox{\linewidth}{!}{%
  \begin{tabular}{ll|cccccccc}
    \toprule
    \multirow{2}{*}{Dataset} & \multirow{2}{*}{Variant} & \multicolumn{8}{c}{Recovery Metrics} \\ 
    \cmidrule{3-10}
    & & FVLoc$\downarrow$ & ActProb$\downarrow$ & Distance$\downarrow$ & Radius$\downarrow$ & SI$\downarrow$ & SD$\downarrow$ & DARD$\downarrow$ & STVD$\downarrow$ \\ 
    \midrule
    \multirow{12}{*}{Tencent} 
    & CAMS-E & \textbf{0.3421} & \underline{0.2444} & 0.2087 & 0.0340 & \textbf{0.2945} & 0.0443 & \textbf{0.2681} & 0.6284 \\
    & CAMS-E w/o C & 0.3761 & 0.2721 & 0.2285 & 0.0688 & 0.3431 & 0.0614 & \underline{0.2814} & 0.6365 \\
    & CAMS-E w/o I & 0.3421 & 0.3721 & 0.1995 & 0.0267 & 0.2985 & 0.0229 & 0.3221 & \textbf{0.6219} \\
    & CAMS-E w/o I\&C & 0.3421 & 0.3887 & \textbf{0.1563} & 0.0660 & 0.3342 & 0.0377 & 0.3166 & 0.6448 \\
    & CAMS-M & 0.3680 & 0.2458 & 0.1838 & \underline{0.0266} & 0.3215 & \underline{0.0334} & 0.3066 & \underline{0.6250} \\
    & CAMS-M w/o C & 0.3421 & 0.3108 & 0.1828 & 0.0790 & \underline{0.2963} & 0.0390 & 0.3271 & 0.6478 \\
    & CAMS-M w/o I & 0.3468 & 0.3248 & 0.1735 & 0.0469 & 0.2722 & \textbf{0.0330} & 0.3150 & 0.6265 \\
    & CAMS-M w/o I\&C & 0.3568 & 0.3794 & 0.1746 & 0.0305 & 0.3147 & 0.0467 & 0.2937 & 0.6283 \\
    & CAMS-S & 0.3421 & \textbf{0.2136} & 0.1699 & \textbf{0.0195} & 0.3790 & 0.0378 & 0.3161 & 0.6301 \\
    & CAMS-S w/o C & 0.3421 & 0.3240 & 0.2058 & 0.0412 & 0.3001 & 0.0435 & 0.3178 & 0.6269 \\
    & CAMS-S w/o I & 0.3421 & 0.3504 & \underline{0.1655} & 0.0596 & 0.2655 & 0.0513 & 0.3115 & 0.6259 \\
    & CAMS-S w/o I\&C & \underline{0.3468} & 0.3518 & 0.1783 & 0.0790 & 0.3202 & 0.0543 & 0.3151 & 0.6356 \\
    \midrule
    
    \multirow{12}{*}{ChinaMobile} 
    & CAMS-E & \textbf{0.3055} & \textbf{0.3664} & \textbf{0.1339} & 0.0717 & 0.3464 & 0.0484 & \textbf{0.3248} & 0.6601 \\
    & CAMS-E w/o C & 0.3055 & 0.3784 & 0.2162 & 0.0975 & 0.3330 & 0.0592 & 0.3393 & 0.6753 \\
    & CAMS-E w/o I & \underline{0.3118} & 0.3771 & 0.1956 & 0.0676 & 0.3207 & \underline{0.0451} & 0.3693 & \textbf{0.6437} \\
    & CAMS-E w/o I\&C & 0.3118 & 0.3742 & 0.2117 & 0.0654 & 0.3500 & 0.0490 & 0.3792 & \underline{0.6501} \\
    & CAMS-M & 0.3055 & 0.3780 & \underline{0.1468} & \textbf{0.0340} & 0.3106 & 0.0633 & 0.3526 & 0.6550 \\
    & CAMS-M w/o C & 0.3118 & 0.3971 & 0.1934 & 0.0704 & 0.3482 & 0.0532 & 0.3592 & 0.6594 \\
    & CAMS-M w/o I & 0.3055 & 0.4141 & 0.2122 & 0.0454 & 0.3198 & \textbf{0.0396} & 0.3574 & 0.6526 \\
    & CAMS-M w/o I\&C & 0.3118 & 0.4100 & 0.1846 & 0.0874 & 0.3643 & 0.0796 & \underline{0.3336} & 0.6546 \\
    & CAMS-S & 0.3055 & \underline{0.3687} & 0.1669 & 0.0601 & \textbf{0.2896} & 0.0578 & 0.3409 & 0.6622 \\
    & CAMS-S w/o C & 0.3055 & 0.3869 & 0.1795 & 0.0775 & \underline{0.2973} & 0.0567 & 0.3427 & 0.6643 \\
    & CAMS-S w/o I & 0.3118 & 0.4006 & 0.1713 & \underline{0.0377} & 0.3081 & 0.0643 & 0.3717 & 0.6572 \\
    & CAMS-S w/o I\&C & 0.3055 & 0.3926 & 0.2303 & 0.0871 & 0.3371 & 0.0708 & 0.3766 & 0.6524 \\
    \bottomrule
  \end{tabular}%
  }
\end{table}

\begin{table}[htbp]
  \centering
  \small
  \caption{Performance comparison of human vs. hierarchical address types in recovery and generation tasks. Best results for each metric are highlighted in \textbf{bold}.}
  \label{tab:address-type-comparison}
  \resizebox{\linewidth}{!}{%
  \begin{tabular}{lll|ccccccccc}
    \toprule
    \multirow{2}{*}{Dataset} & \multirow{2}{*}{Task} & \multirow{2}{*}{Address Type} & \multicolumn{9}{c}{Performance Metrics} \\ 
    \cmidrule{4-12}
    & & & Loc$\downarrow$ & Prob$\downarrow$ & Dist$\downarrow$ & Rad$\downarrow$ & SI$\downarrow$ & DailyLoc$\downarrow$ & SD$\downarrow$ & DARD$\downarrow$ & STVD$\downarrow$ \\ 
    \midrule
    \multirow{4}{*}{Tencent} 
    & \multirow{2}{*}{Recovery} & Human & \textbf{0.3421} & \textbf{0.3383} & \textbf{0.2728} & \textbf{0.1212} & \textbf{0.4479} & 0.1803 & \textbf{0.0578} & \textbf{0.3849} & \textbf{0.6391} \\
    & & Hierarchical & 0.3420 & 0.3117 & 0.2852 & 0.1004 & 0.4639 & \textbf{0.1397} & 0.0779 & 0.4099 & 0.6336 \\
    \cmidrule{2-12}
    & \multirow{2}{*}{Generation} & Human & \textbf{0.3148} & \textbf{0.3611} & \textbf{0.3687} & 0.1533 & \textbf{0.4566} & \textbf{0.1687} & \textbf{0.0845} & \textbf{0.3931} & \textbf{0.6533} \\
    & & Hierarchical & 0.3089 & 0.3840 & 0.4577 & \textbf{0.1283} & 0.4690 & 0.1975 & 0.1114 & 0.4092 & 0.6512 \\
    \midrule
    
    \multirow{4}{*}{ChinaMobile} 
    & \multirow{2}{*}{Recovery} & Human & 0.3118 & \textbf{0.3872} & 0.2942 & \textbf{0.0691} & \textbf{0.3054} & 0.1217 & 0.0619 & \textbf{0.3935} & \textbf{0.6400} \\
    & & Hierarchical & 0.3118 & 0.4648 & \textbf{0.2828} & 0.0986 & 0.4945 & \textbf{0.0379} & \textbf{0.0587} & 0.4143 & 0.6425 \\
    \cmidrule{2-12}
    & \multirow{2}{*}{Generation} & Human & 0.3059 & \textbf{0.3317} & 0.3807 & 0.0778 & \textbf{0.4624} & \textbf{0.0980} & \textbf{0.0767} & \textbf{0.3757} & 0.6591 \\
    & & Hierarchical & \textbf{0.2992} & 0.3586 & \textbf{0.3251} & \textbf{0.0767} & 0.4789 & 0.1497 & 0.0909 & 0.3888 & \textbf{0.6614} \\
    \bottomrule
  \end{tabular}%
  }
\end{table}
  
The hierarchical address representation is designed to activate geographic knowledge in CityGPT by associating location names with their attributes across defined regional hierarchies, resulting in generating more accurate locations with reduced hallucinations. In comparison, the human-centric address representation directly prompt the model to recall neighborhood geographic information from the training corpus. Results in Table \ref{tab:address-type-comparison} demonstrate that the second one performs better, likely because training corpus of CityGPT has been pre-aligned with OSM data.

\subsection{Generated mobility patterns of different user profiles }\label{user mobility patterns}
\begin{figure*}
  \centering
  \begin{minipage}[b]{0.55\textwidth}
    \centering
    \newcommand{\subfigwidth}{0.31\linewidth} %
    \newcommand{\subfigheight}{1.7cm}
    \newcommand{\hspacing}{0.01cm} %
    
    \begin{subfigure}[t]{\subfigwidth}
      \includegraphics[width=\linewidth, height=\subfigheight]{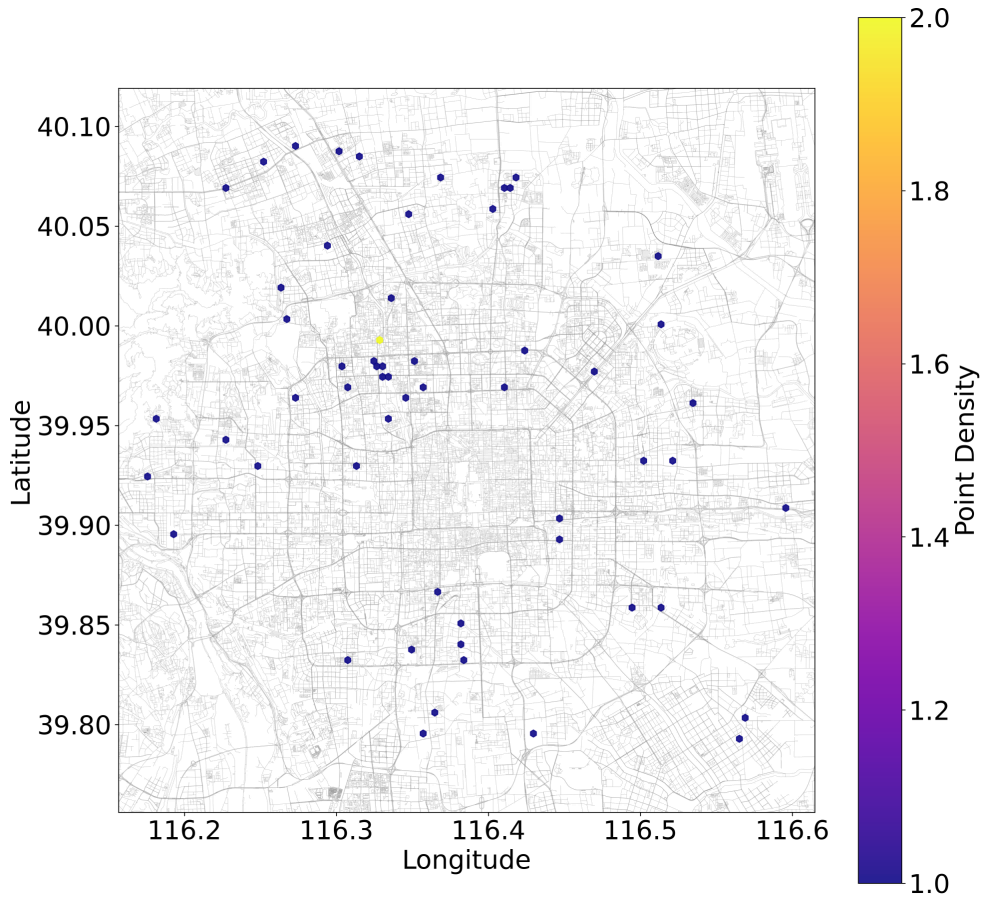}
    \end{subfigure}
    \hspace{\hspacing}
    \begin{subfigure}[t]{\subfigwidth}
      \includegraphics[width=\linewidth, height=\subfigheight]{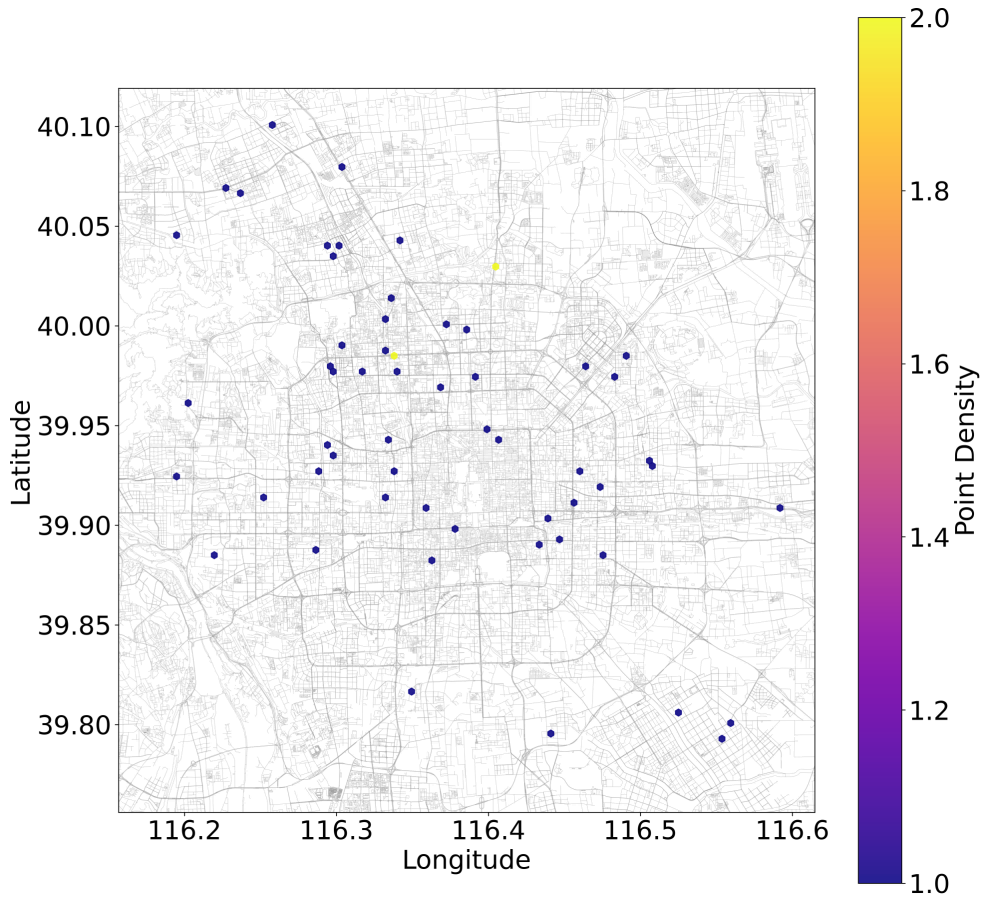}
    \end{subfigure}
    \hspace{\hspacing}
    \begin{subfigure}[t]{\subfigwidth}
      \includegraphics[width=\linewidth, height=\subfigheight]{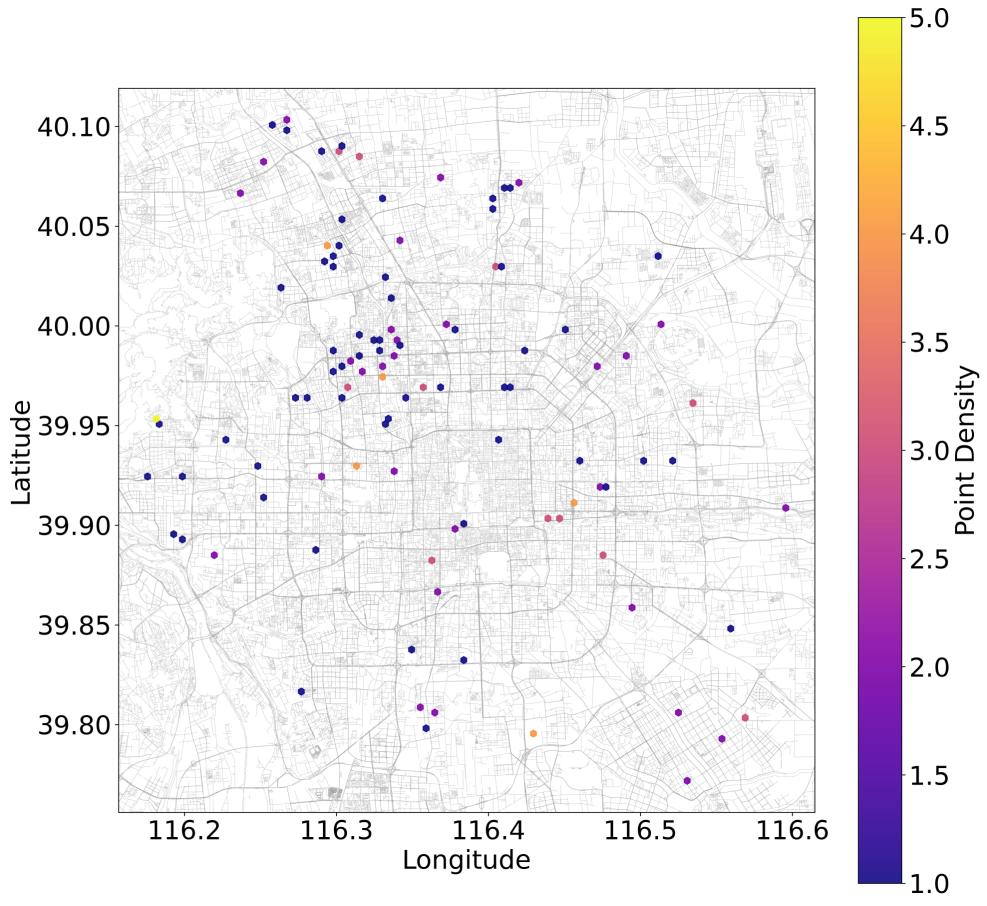}
    \end{subfigure}
    
    \vspace{0.2cm}
    
    \begin{subfigure}[t]{\subfigwidth}
      \includegraphics[width=\linewidth, height=\subfigheight]{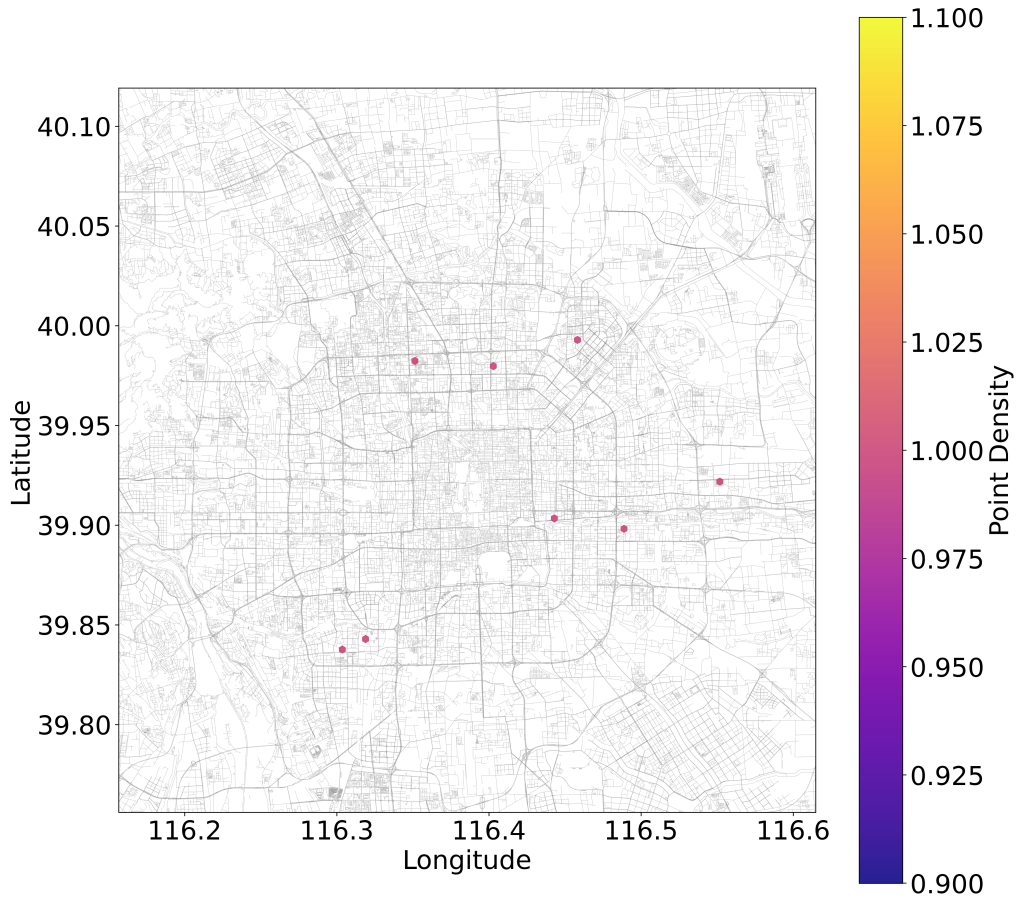}
    \end{subfigure}
    \hspace{\hspacing}
    \begin{subfigure}[t]{\subfigwidth}
      \includegraphics[width=\linewidth, height=\subfigheight]{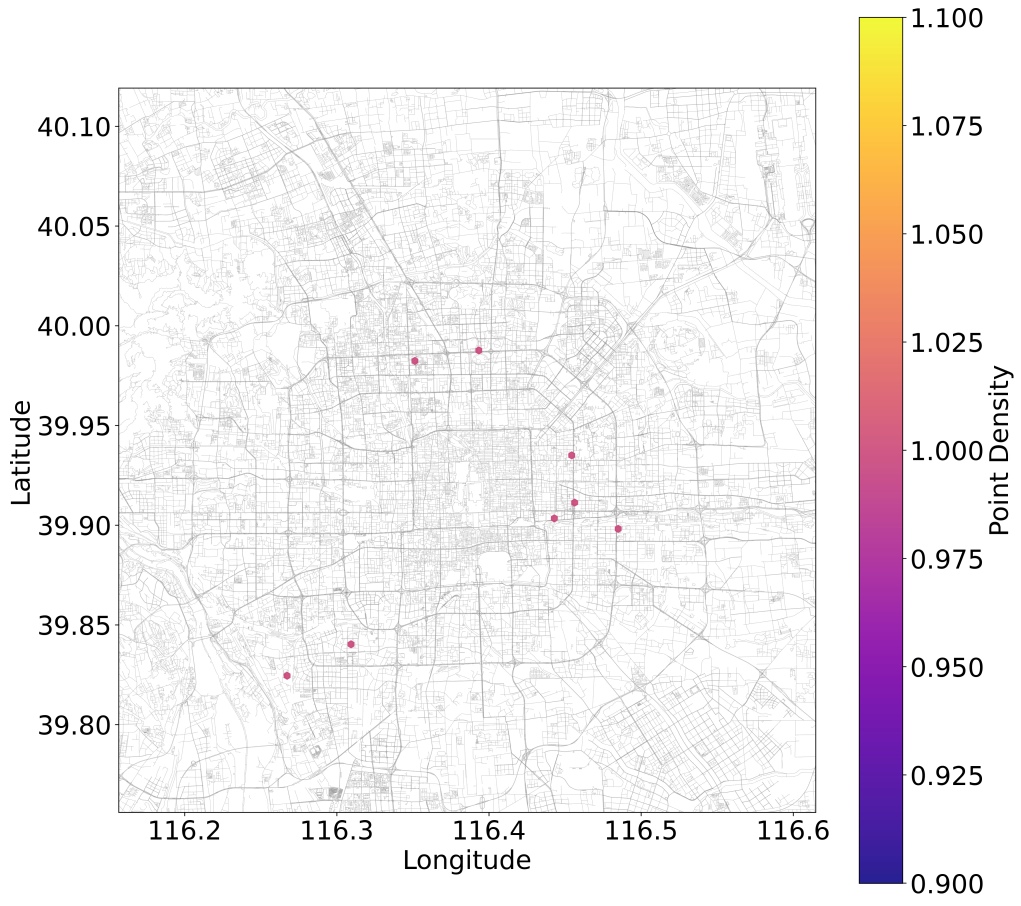}
    \end{subfigure}
    \hspace{\hspacing}
    \begin{subfigure}[t]{\subfigwidth}
      \includegraphics[width=\linewidth, height=\subfigheight]{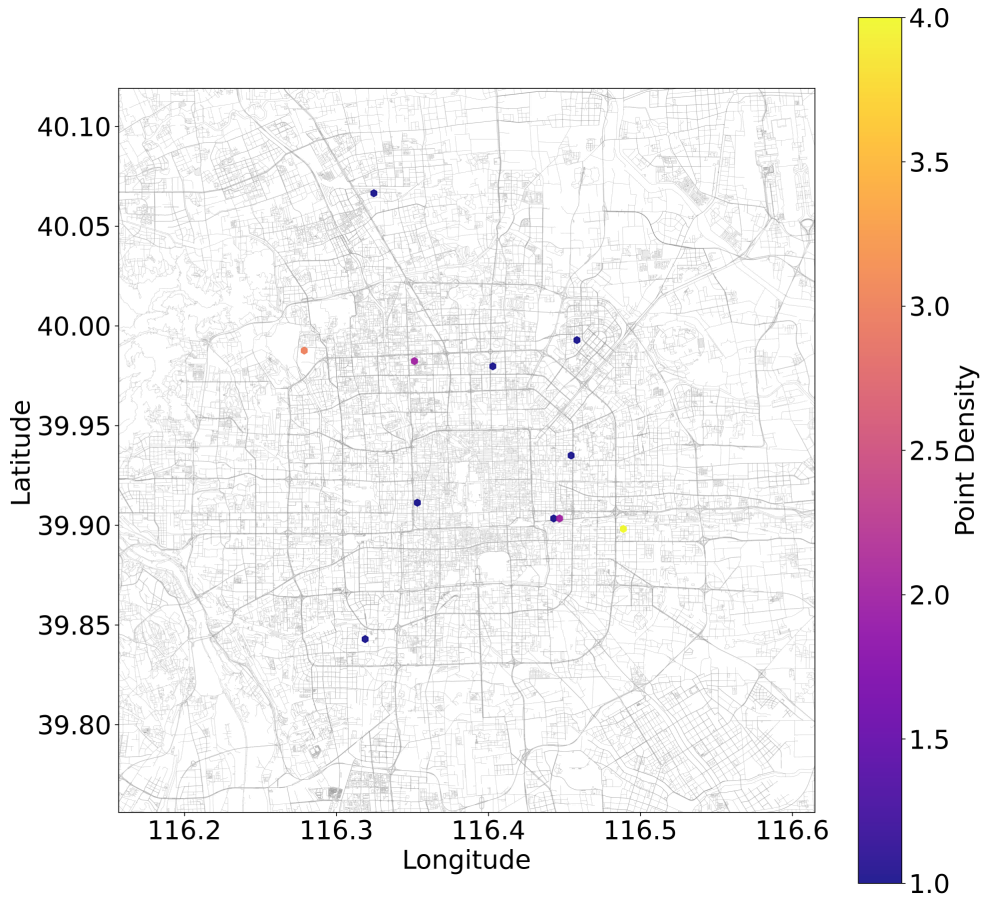}
    \end{subfigure}
    
    \vspace{0.2cm}
    
    \begin{subfigure}[t]{\subfigwidth}
      \includegraphics[width=\linewidth, height=\subfigheight]{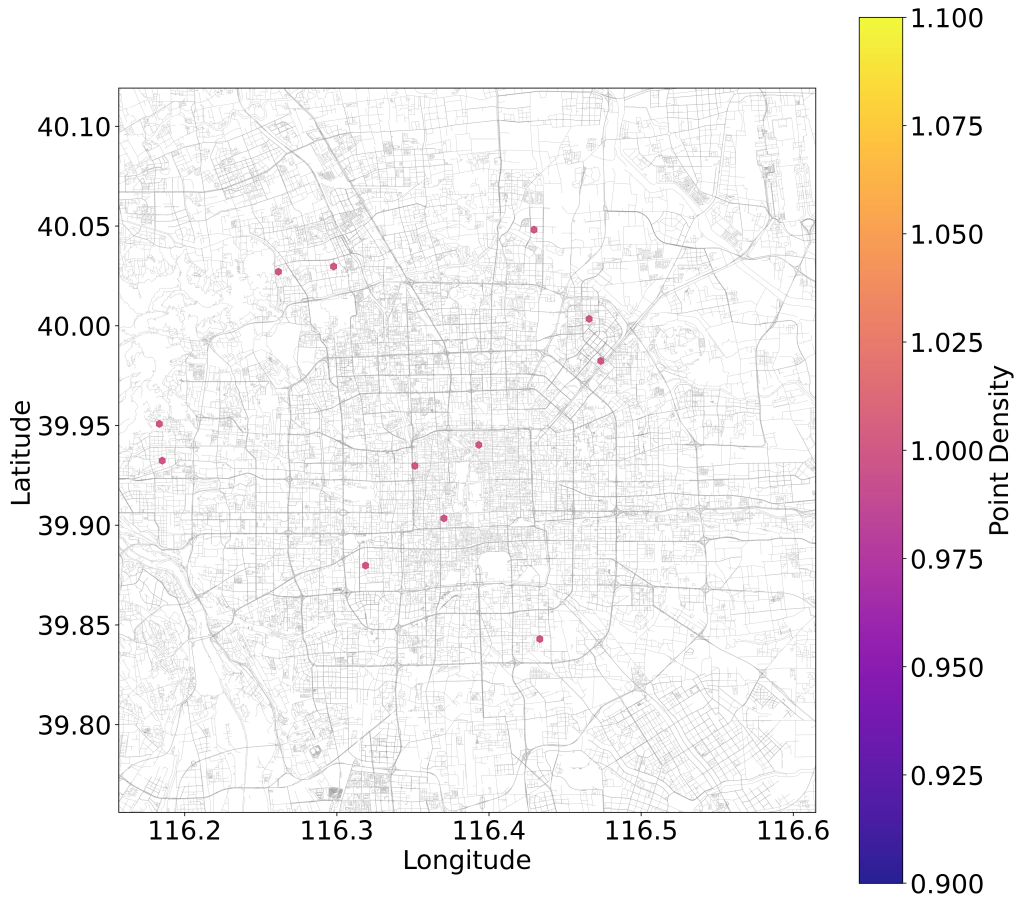}
      \caption{Home}
      \label{fig:home-col}
    \end{subfigure}
    \hspace{\hspacing}
    \begin{subfigure}[t]{\subfigwidth}
      \includegraphics[width=\linewidth, height=\subfigheight]{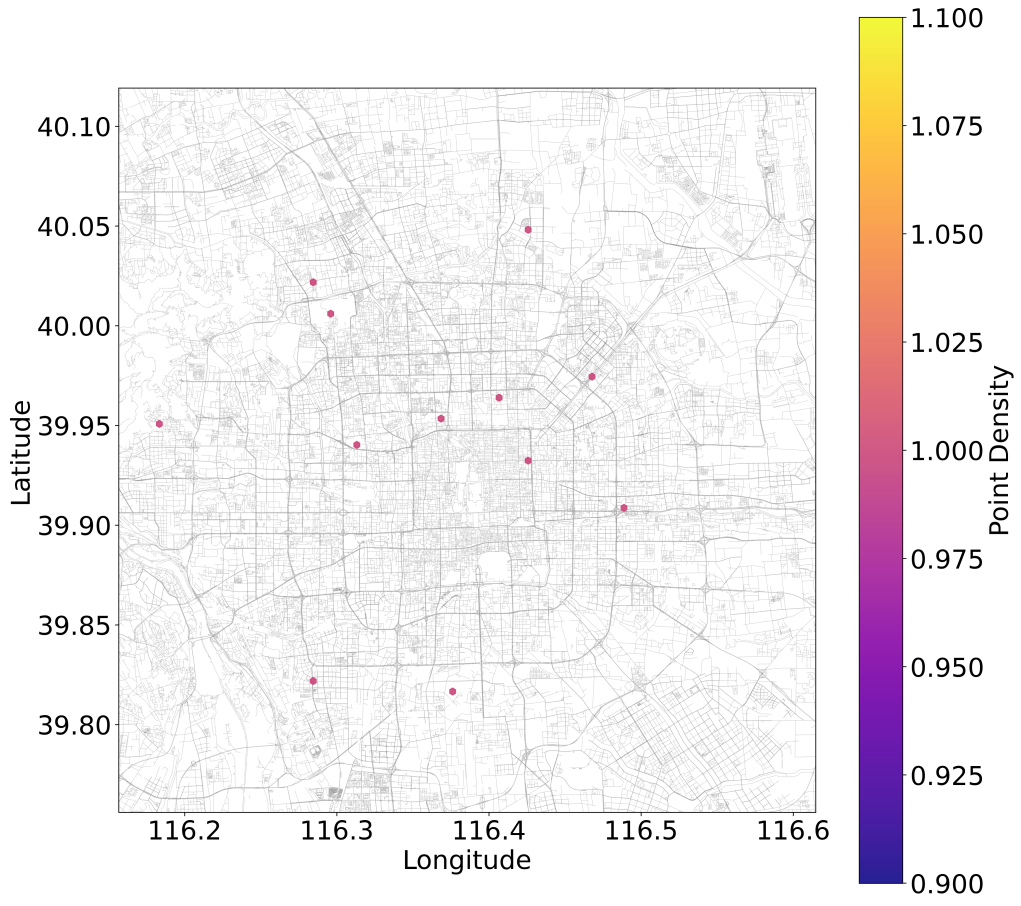}
      \caption{Workplace}
      \label{fig:work-col}
    \end{subfigure}
    \hspace{\hspacing}
    \begin{subfigure}[t]{\subfigwidth}
      \includegraphics[width=\linewidth, height=\subfigheight]{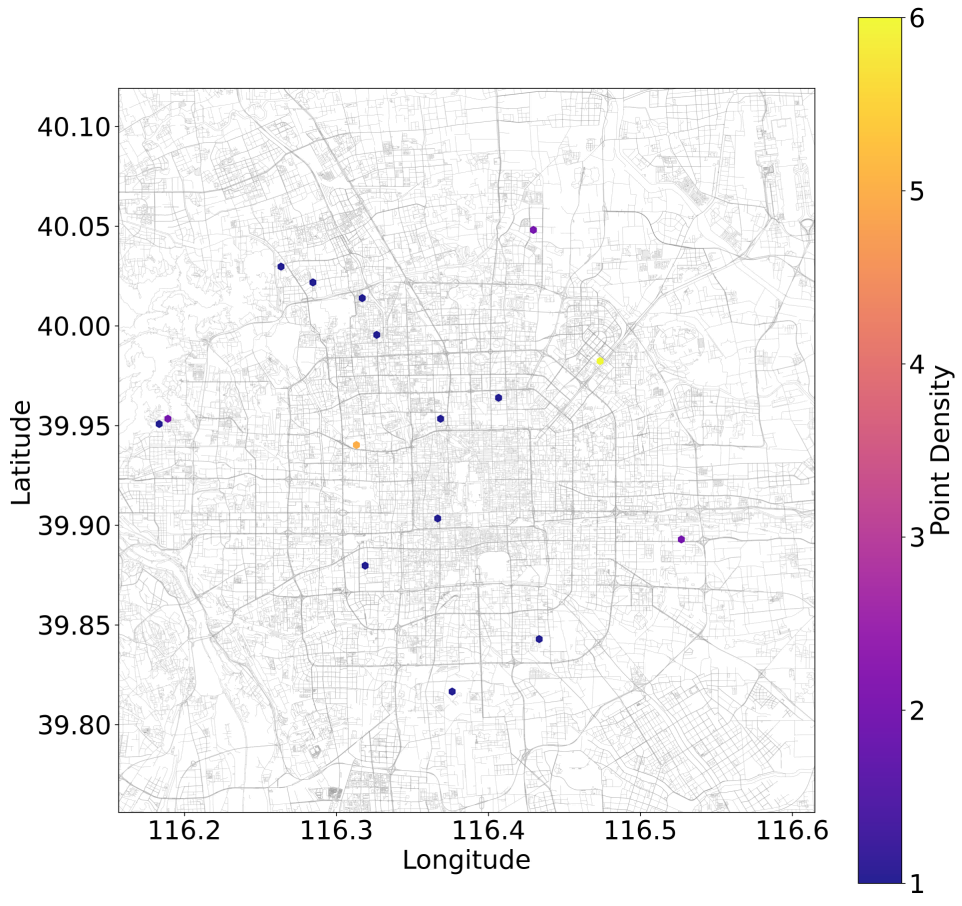}
      \caption{Trajectory}
      \label{fig:traj-col}
    \end{subfigure}
  \end{minipage}
  \hspace{0.1cm} %
  \begin{minipage}[b]{0.4\textwidth}
    \centering
    \includegraphics[width=\linewidth, height=5.1cm]{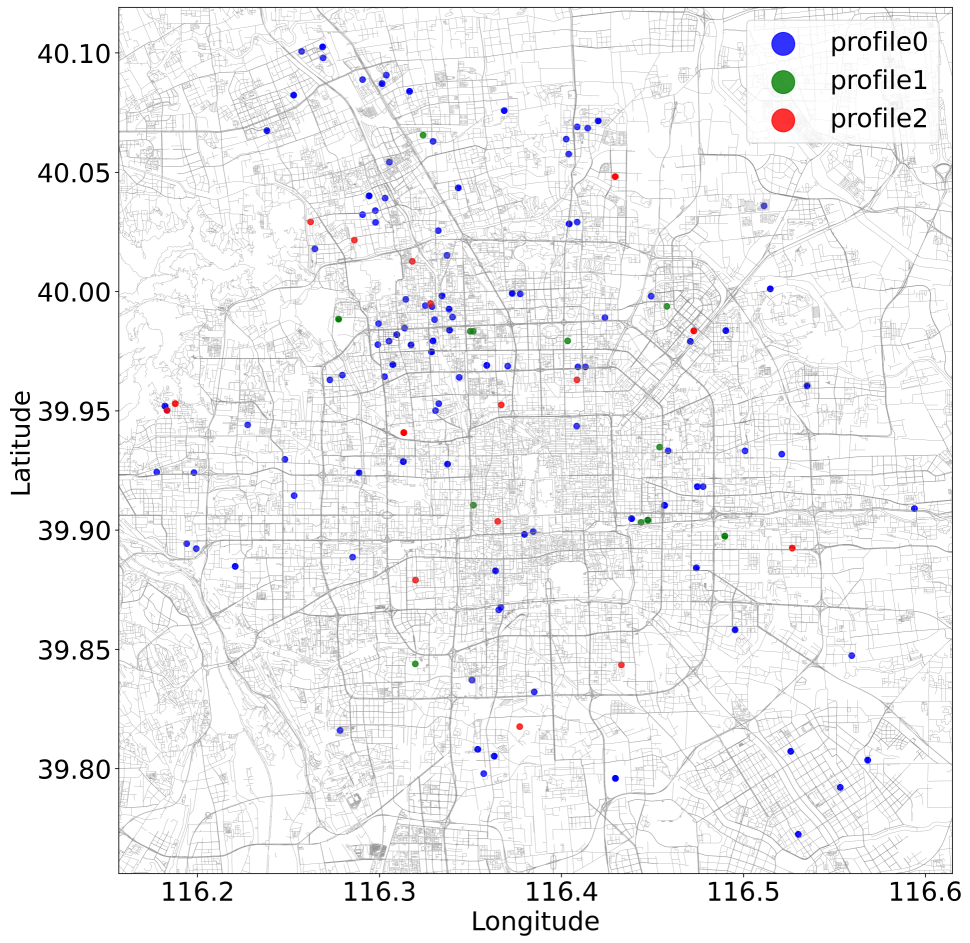}
    \caption{Comparison of daily movement patterns}
    \label{fig:comparative-map}
  \end{minipage}
  
  \caption{Comparison of Mobility patterns in real urban spaces of different user profiles.}
  \label{fig:spatial-patterns}
\end{figure*}

\begin{table}[htbp]
  \centering
  \small
  \caption{Performance comparison of embedding-based vs. LLM-based similarity methods in trajectory generation. Best results for each metric are highlighted in \textbf{bold}.}
  \label{tab:similar-methods-comparison}
  \resizebox{\linewidth}{!}{%
  \begin{tabular}{ll|ccccccccc}
    \toprule
    \multirow{2}{*}{Dataset} & \multirow{2}{*}{Similarity Method} & \multicolumn{8}{c}{Generation Metrics} \\ 
    \cmidrule{3-10}
    & & Loc$\downarrow$ & Prob$\downarrow$ & Dist$\downarrow$ & Rad$\downarrow$ & SI$\downarrow$ & SD$\downarrow$ & DARD$\downarrow$ & STVD$\downarrow$ \\ 
    \midrule
    \multirow{2}{*}{Tencent} 
    & Embedding & 0.3148 & 0.3611 & \textbf{0.3687} & \textbf{0.1533} & \textbf{0.4566} & 0.0845 & \textbf{0.3931} & \textbf{0.6533} \\
    & LLM & \textbf{0.3027} & \textbf{0.3474} & 0.3950 & 0.1674 & 0.4679 & \textbf{0.0650} & 0.3936 & 0.6583 \\
    \midrule
    
    \multirow{2}{*}{ChinaMobile} 
    & Embedding & 0.3118 & \textbf{0.3872} & \textbf{0.2942} & \textbf{0.0691} & 0.5054 & 0.0619 & \textbf{0.3935} & \textbf{0.6400} \\
    & LLM & \textbf{0.2992} & 0.4792 & 0.3701 & 0.0794 & \textbf{0.4766} & \textbf{0.0616} & 0.4156 & 0.6546 \\
    \bottomrule
  \end{tabular}%
  }
\end{table}

We compared the distributions of anchor points and trajectory points generated by the model for three types of user profiles in Figure  \ref{fig:spatial-patterns}. The first row in the left panel represents to the user profiles of 20-year-old, moderately-income male with a bachelor's degree working as an IT engineer(Profile 0). Generated work locations (Figure \ref{fig:work-col})of this group show a higher degree of clustering in city centers compared to their residential locations(Figure \ref{fig:home-col}), and the overall trajectory point distribution is relatively dispersed(Figure \ref{fig:traj-col}), suggesting that they tend to reside in suburban areas to reduce living costs, while working in IT-related companies concentrated in urban centers. The second row corresponds to young females in Finance/Accounting/Auditing/Tax/Cashier with a bachelor's degree and a moderate income(Profile 1). Generated workplaces and residences of Profile 1 cluster around major corporate hubs, suggesting that they often work in larger companies and rent apartments nearby their workplaces. The third row represents middle-aged males in Network Sales/Operations/Services with a bachelor's degree and a moderate income(Profile 2).Compared to the previous two user types, the trajectories generated for Profile 2 appear more scattered and irregular, demonstrating less regular mobility patterns, as they may travel more frequently for business.

\end{document}